\RequirePackage{fix-cm}
\documentclass[smallcondensed]{svjour3}     
\smartqed  
\usepackage{graphicx}
\usepackage{natbib}
\bibpunct{(}{)}{;}{a}{}{,}
\usepackage{amsmath}
\usepackage{amsmath, amssymb, latexsym}
\usepackage{commath}
\usepackage{tabularx}
\usepackage{rotating}

\DeclareMathOperator{\E}{\mathbb{E}}
\DeclareMathOperator{\PP}{\mathbb{P}}

\begin{document}

\title{A framework for the fine-grained evaluation of the instantaneous expected value of soccer possessions}

\titlerunning{A framework for the evaluation of the expected value of soccer possessions}        

\author{Javier Fern\'andez \and
        Luke Bornn \and 
        Daniel Cervone
}



\institute{Javier Fern\'andez \at
              FC Barcelona and Polytechnic University of Catalonia, Barcelona, Spain.\\
              \email{ javier.fernandezr@fcbarcelona.cat / javier.fernandez.de.la.rosa@upc.edu}           
           \and
           Luke Bornn \at
           Simon Fraser University
           \and
           Dan Cervone \at
           Zelus Analytics
}

\date{Received: date / Accepted: date}

\maketitle

\begin{abstract}

The expected possession value (EPV) of a soccer possession represents the likelihood of a team scoring or receiving the next goal at any time instance. By decomposing the EPV into a series of subcomponents that are estimated separately, we develop a comprehensive analysis framework providing soccer practitioners with the ability to evaluate the impact of both observed and potential actions. We show we can obtain calibrated models for all the components of EPV, including a set of yet-unexplored problems in soccer. We produce visually-interpretable probability surfaces for potential passes from a series of deep neural network architectures that learn from low-level spatiotemporal data. Additionally, we present a series of novel practical applications providing coaches with an enriched interpretation of specific game situations.

\end{abstract}
\section{Introduction}

Professional sports teams have started to gain a competitive advantage in recent decades by using advanced data analysis. However, soccer has been a late bloomer in integrating analytics, mainly due to the difficulty of making sense of the game's complex spatiotemporal relationships. To address the nonstop flow of questions that coaching staff deal with daily, we require a flexible framework that can capture the complex spatial and contextual factors that rule the game while providing practical interpretations of real game situations. This paper addresses the problem of estimating the expected value of soccer possessions (EPV) and proposes a decomposed learning approach that allows us to obtain fine-grained visual interpretations from neural network-based components.\\

\begin{figure*}
  \includegraphics[width=1\textwidth]{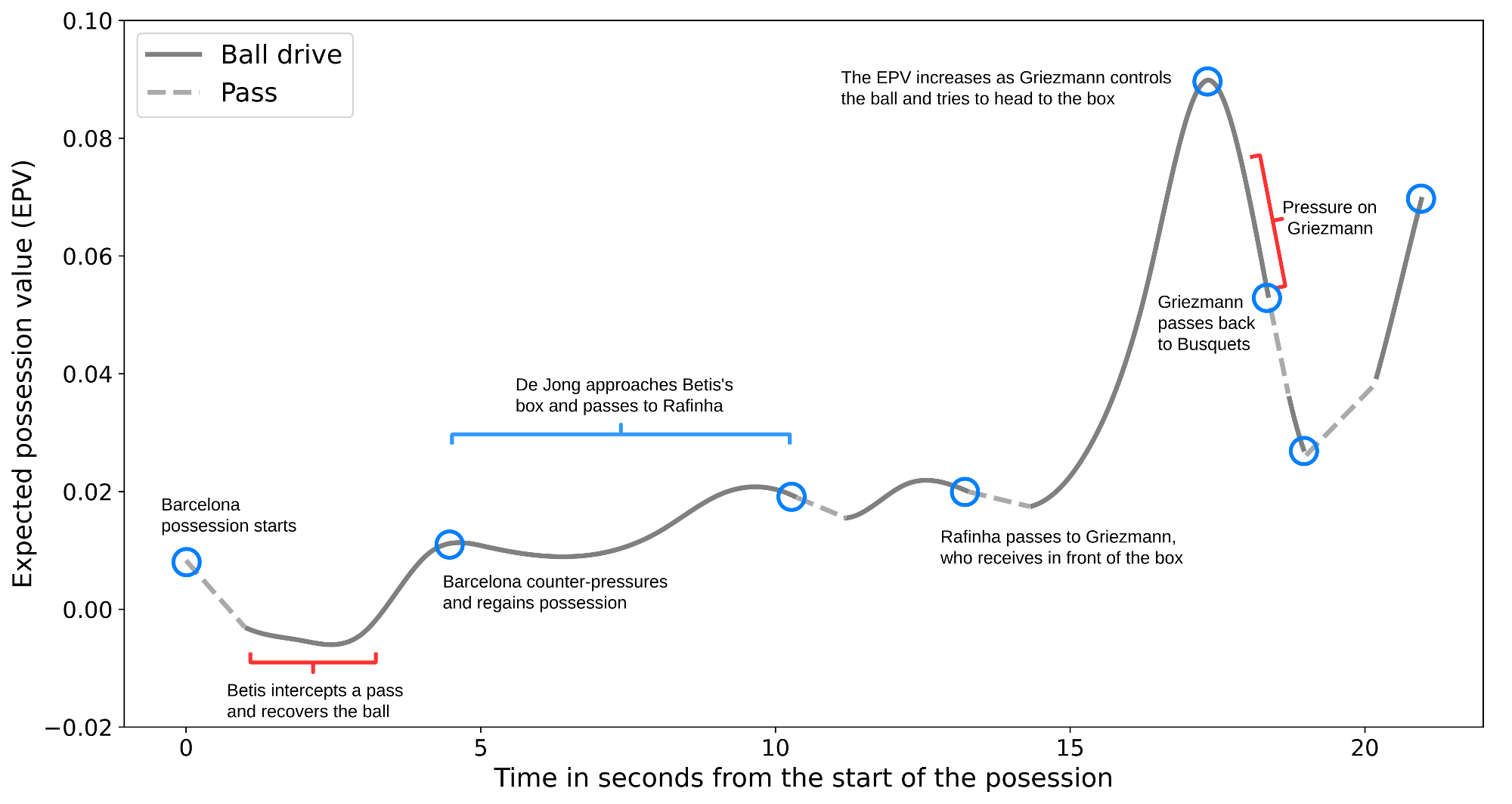}
\caption{Evolution of the expected possession value (EPV) of FC Barcelona during a match against Real Betis in La Liga season 2019/2020.}
\label{fig:epv}
\end{figure*}

The EPV is essentially an estimate of which team will score the next goal, given all the spatiotemporal information available at any given time. Let $G \in \{-1,1\}$, where the values represent one or the other team scoring next, respectively; the EPV corresponds to the expected value of $G$. The frame-by-frame estimation of EPV constitutes a one-dimensional time series that provides an intuitive description of how the possession value changes in time, as presented in Figure \ref{fig:epv}. While this value alone can provide precise information about the impact of observed actions, it does not provide sufficient practical insight into either the factors that make it fluctuate or which other advantageous actions could be taken to boost EPV further. To reach this granularity level, we formulate EPV as a composition of the expectations of three different on-ball actions: passes, ball drives, and shots. Each of these components is estimated separately, producing an ensemble of models whose outputs can be merged to produce a single EPV estimate. Additionally, by inspecting each model, we can obtain detailed insight on the impact that each of the components has on the final EPV estimation.\\

We propose two different approaches to learn each of the separated models, depending on whether we need to consider each possible location on the field or just single locations. We propose several deep neural architectures capable of producing full prediction surfaces from low-level features for the first case. We show that it is possible to learn these surfaces from very challenging learning set-ups where only a single-location ground-truth correspondence is available for estimating the whole surface. For the second case, we use shallow neural networks on top of a broad set of novel spatial and contextual features. From a practical standpoint, we are splitting out a complex model into more easily understandable parts so the practitioner can both understand the factors that produce the final estimate and evaluate the effect that other possible actions may have had. This type of modeling allows for easier integration of complex prediction models into the decision-making process of groups of individuals with a non-scientific background. Also, each of the components can be used individually, multiplying the number of potential applications. 


The main contributions of this work are the following:
\begin{itemize}
\item We propose a framework for estimating the instantaneous expected outcome of any soccer possession, which allows us to provide professional soccer coaches with rich numerical and visual performance metrics. 
\item We show that by decomposing the target EPV expression into a series of sub-components and estimating these separately, we can obtain accurate and calibrated estimates and provide a framework with greater interpretability than single-model approaches  \citep{cervone2016multiresolution,bransen2018measuring}.
\item We develop a series of deep learning architectures to estimate the expected possession value surface of potential passes, pass success probability, pass selection probability surfaces, and show these three networks provide both accurate and calibrated surface estimates.
\item We present a handful of novel practical applications in soccer that are directly derived from this framework.
\end{itemize}

\section{Background}

The evaluation of individual actions has been recently gaining attention in soccer analytics research. Given the relatively low frequency of soccer goals compared to match duration and the frequency of other events such as passes and turnovers, it becomes challenging to evaluate individual actions within a match. Several different approaches have been attempted to learn a valuation function for both on-ball and off-ball events related to goal-scoring.\\

Handcrafted features based on the opinion of a committee of soccer experts have been used to quantify the likelihood of scoring in a continuous-time range during a match \citep{link2016real}. Another approach uses observed events' locations to estimate the value of individual actions during the development of possessions \citep{decroos2018actions}. Here, the game state is represented as a finite set of consecutive observed discrete actions and, a Bernoulli distributed outcome variable is estimated through standard supervised machine learning algorithms. In a similar approach, possession sequences are clustered based on dynamic time warping distance, and an XGBoost \citep{chen2016xgboost} model is trained to predict the expected goal value of the sequence, assuming it ends with a shot attempt \citep{bransen2018measuring}. 
 \cite{gyarmati2016qpass}, calculate the value of a pass as the difference of field value between different locations when a ball transition between these occurs.  \cite{rudd2011framework} uses Markov chains to estimate the expected possession value based on individual on-ball actions and a discrete transition matrix of 39 states, including zonal location, defensive state, set pieces, and two absorbing states (goal or end of possession). A similar approach named expected threat uses Markov chains and a coarsened representation of field locations to derive the expected goal value of transitioning between discrete locations \citep{xtkarun}.
The estimation of a shot's expectation within the next 10 seconds of a given pass event has also been used to estimate a pass's reward, based on spatial and contextual information \citep{power2017not}. 
Beyond the quantification of on-ball actions, off-ball position quality has also been quantified, based on the goal expectation. In \cite{spearmanbeyond}, a physics-based statistical model is designed to quantify the quality of players' off-ball positioning based on the positional characteristics at the time of the action that precedes a goal-scoring opportunity. All of these previous attempts on quantifying action value in soccer assume a series of constraints that reduce the scope and reach of the solution. Some of the limitations of these past work include simplified representations of event data (consisting of merely the location and time of on-ball actions), using strongly handcrafted rule-based systems, or focusing exclusively on one specific type of action. However, a comprehensive EPV framework that considers both the full spatial extent of the soccer field and the space-time dynamics of the 22 players and the ball has not yet been proposed and fully validated. In this work, we provide such a framework
and go one step further estimating the added value of observed actions by providing an approach for estimating the expected value of the possession at any time instance.\\

Action evaluation has also been approached in other sports such as basketball and ice-hockey by using spatiotemporal data. The expected possession value of basketball possessions was estimated through a multiresolution process combining macro-transitions (transitions between states following coarsened representation of the game state) and micro-transitions (likelihood of player-level actions), capturing the variations between actions, players, and court space \citep{cervone2016multiresolution}. Also, deep reinforcement learning has been used for estimating an action-value function from event data of professional ice-hockey games \citep{liu2018deep}. Here, a long short-term memory deep network is trained to capture complex time-dependent contextual features from a set of low-level input information extracted from consecutive on-puck events.


\section{Structured modeling}

In this study, we aim to provide a model for estimating soccer possessions' expected outcomes at any given time. While the single EPV estimate has practical value itself, we propose a structured modeling approach where the EPV is decomposed into a series of subcomponents. Each of these components can be estimated separately,  providing the model with greater adaptability to component-specific problems and facilitating the final estimate's interpretation.

\subsection{EPV as a Markov decision process}\label{sec:epv_markov}
This problem can be framed as a Markov decision process (MDP). Let a player with possession of the ball be an agent that can take any action of a discrete set $A$ from any state of the set of all possible states $S$; we aim to learn the state-value function $EPV(s)$, defined as the expected return from state $s$, based on a policy $\pi(s,a)$, which defines the probability of taking action $a$ at state $s$. In contrast with typical MDP applications, our aim is not to find the optimal policy $\pi$, but to estimate the expected possession value (EPV) from an average policy learned from historical data.\\

Let $\Gamma$ be the set of all possible soccer possessions, and  $r \in \Gamma$ represents the full path of a specific possession. Let $\Psi$ be a high dimensional space, including all the spatiotemporal information and a series of annotated events, $T_t(r) \in \Psi$ is a snapshot of the spatiotemporal data after $t$ seconds from the start of the possession. And let $G(r)$ be the outcome of a possession $w$, where $G(r) \in \{-1,1\}$, with $1$ being a goal is scored and $-1$  being a goal is conceded.\\

\begin{definition}\label{def:epv_general}
The expected possession value of a soccer possession at time $t$ is $EPV_t = \E[G| T_t]$
\end{definition}

This initial definition shares similarities with previous approaches in other sports, such as basketball \citep{cervone2016multiresolution} and American football \citep{yurko2019going}, from which part of the notation used in this section is inspired. Following Definition \ref{def:epv_general}, we can observe that EPV is an integration over all the future paths a possession can take at time $t$, given the available spatiotemporal information at that time, $T_t$. We employ player tracking data consisting of the location of the 22 players and the ball, usually provided at a frequency ranging from 10Hz to 25Hz, and captured using computer-vision algorithms on top of videos of professional soccer matches. We will assume that tracking data is accompanied and synchronized with event data, consisting of annotated events observed during the match, indicating the location, time, and other possible tags. Let $\Psi$ be the infinite set of possible tracking data snapshots; this modeling approach defines a continuous state space, represented by $\Psi$.\\


\subsection{A decomposed model}\label{sec:decomposed}

In order to obtain the desired structured modeling of EPV described in Section \ref{sec:epv_markov}, we will further decompose Definition \ref{def:epv_general} following the law of total expectation and considering the set of possible actions that can be taken at any given time. We assume that the space of possible actions $A=\{\rho, \delta,\varsigma\}$ is a discrete set where $\rho$, $\delta$, and $\varsigma$ represent pass, ball drive, and shot attempt actions, respectively. We can rewrite Definition \ref{def:epv_general} as in Equation \ref{eq:structured_epv}.

\begin{equation}\label{eq:structured_epv}
\begin{split}
    EPV_t =\sum_{a \in A}\E[G| A=a, T_t]\overbrace{\PP(A=a | T_t)}^{\parbox{7em}{\footnotesize\centering Action selection probability}}
\end{split}
\end{equation}


Additionally, to consider that passes can go anywhere on the field, we define $D_t$ to be the selected pass destination location at time $t$ and $\PP(D_t| T_t)$ to be a transition probability model for passes.  Let $L$ be the set of all the possible locations in a soccer field, then $D_t \in L$. On the other hand, we assume that ball drives ($\delta$) and shots ($\varsigma$) have a single possible destination location (the expected player location in one second and the goal line center, respectively). Following this, we can rewrite Definition \ref{def:epv_general} as presented in Equation \ref{eq:structured_epv}.

\begin{equation}\label{eq:structured_epv}
\begin{split}
    EPV_t =(\sum_{l \in L} \overbrace{\E[G| A=\rho, D_t = l, T_t]}^{\parbox{8em}{\footnotesize\centering Joint expected value surface of passes}} \overbrace{\PP(D_t=l |A=\rho,  T_t)}^{\parbox{8em}{\footnotesize\centering Pass selection probability}}) \PP(A=\rho | T_t) \\ 
     + \overbrace{\E[G| A=\delta, T_t]}^{\parbox{6em}{\footnotesize\centering Expected value of ball drives}} \PP(A=\delta | T_t) \\
     + \overbrace{\E[G| A=\varsigma, T_t]}^{\parbox{6em}{\footnotesize\centering Expected value from shots}} \PP(A=\varsigma | T_t)
\end{split}
\end{equation}


The expected value of passing actions, $\E[G| D, A=\rho]$, can be further extended to include the two scenarios of producing a successful or a missed pass (turnover). We model the outcome of a pass as $O_{\rho}$, which takes a value of $1$ when a pass is successful or $0$ in case of a turnover. We can then rewrite this expression as in Equation \ref{eq:pass_expectation}.

\begin{equation}\label{eq:pass_expectation}
\begin{split}
    \E[G| A=\rho, D_t, T_t]  = \overbrace{\E[G| A=\rho, O_{\rho}=1, D_t, T_t]}^{\parbox{11em}{\footnotesize\centering Expected value of successful/missed passes}} \overbrace{\PP(O_{\rho}=1 | A=\rho, D_t, T_t)}^{\parbox{9em}{\footnotesize\centering Probability of successful/missed passes}}\\
    + \E[G| A=\rho, O_{\rho}=0, D_t, T_t]\PP(O_{\rho}=0 | A=\rho, D_t, T_t)
\end{split}
\end{equation}

Equation \ref{eq:ball_drive_expectation} represents an analogous, definition for ball drives, having $O_{\delta}$ be a random variable taking values $0$ or $1$, representing a successful ball drive or a loss of possession following that ball drive, which we will refer as a missed ball drive.

\begin{equation}\label{eq:ball_drive_expectation}
\begin{split}
    \E[G| A=\delta,  T_t]  = \overbrace{\E[G| A=\delta, O_{\delta}=1,  T_t]}^{\parbox{8em}{\footnotesize\centering Expected value of successful/missed ball drives}} \overbrace{\PP(O_{\delta}=1 | A=\delta,  T_t)}^{\parbox{8em}{\footnotesize\centering Probability of successful/missed ball drives}}\\
    + \E[G| A=\delta, O_{\delta}=0,  T_t]\PP(O_{\delta}=0 | A=\delta,  T_t)
\end{split}
\end{equation}

\begin{figure*}
  \includegraphics[width=1\textwidth]{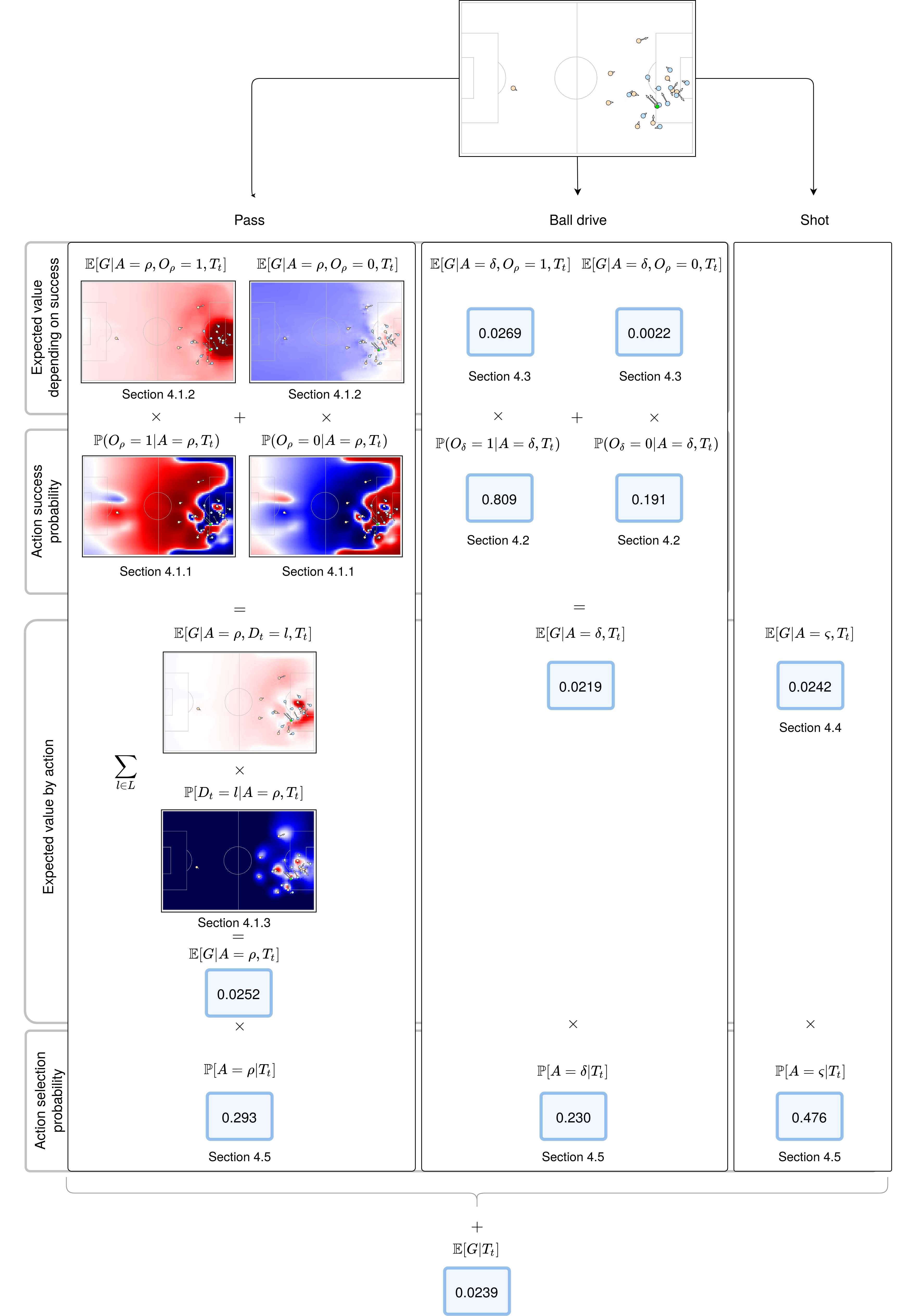}
\caption{Diagram representing the estimation of the expected possession value (EPV) for a given game situation through the composition of independently trained models. The final EPV estimation of $0.0239$ is produced by combining the expected value of three possible actions the player in possession of the ball can take (pass, ball drive, and shot) weighted by the likelihood of those actions being selected. Both pass expectation and probability are modeled to consider every possible location of the field as a destination; thus the diagram presents the predicted surfaces for both successful and unsuccessful potential passes, as well as the surface of destination location likelihood.}
\label{fig:all_layers_epv}
\end{figure*}

Finally, the expression $\E[G|A=\varsigma]$ is equivalent to an expected goals model, a popular metric in soccer analytics \citep{lucey2014quality,eggels2016expected} which models the expectation of scoring a goal based on shot attempts. In Figure \ref{fig:all_layers_epv} we present how the outputs of the different components presented in this section are combined to produce a single EPV estimation, while also providing numerical and visual information of how each part of the model impacts the final value.

\section{Spatiotemporal feature extraction}\label{sec:feature_extraction}

\begin{figure*}[h!]
  \includegraphics[width=1\textwidth]{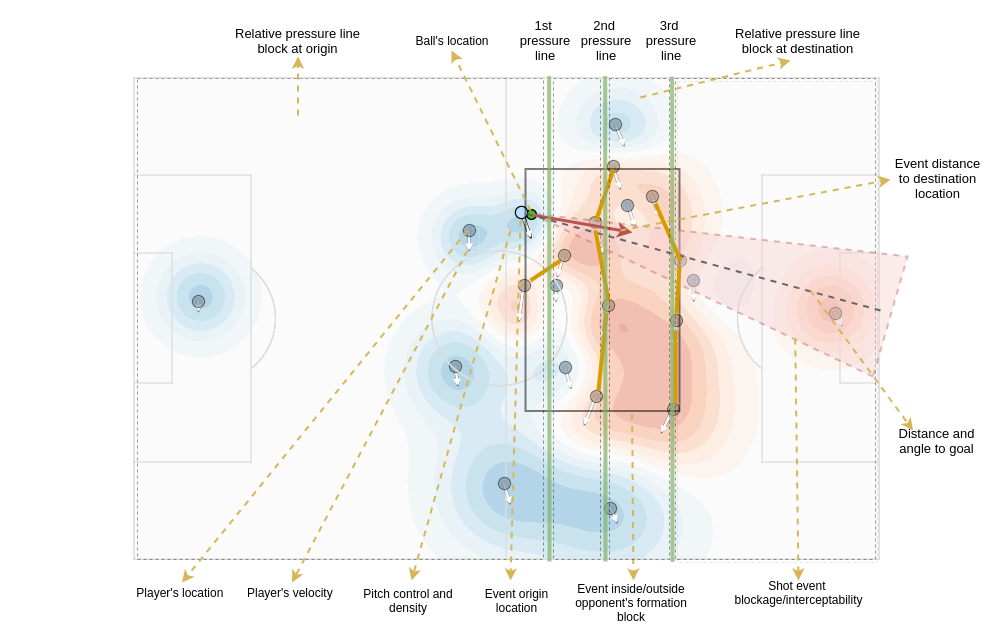}
\caption{Visual representation of a tracking data snapshot of spatial and contextual features in a soccer match situation. Yellow and blue shaded dots represent players of the attacking and defending team, respectively, while the green dot represents the ball location. The red and blue surface represents the pitch control of the attacking team along the field. The grey rectangle covering the yellow dots represents the opponent's formation block. The green vertical lines represent the defending team's vertical dynamic pressure lines, while the polygons with solid yellow lines represent the players clustered in each pressure line. The black dotted rectangles represent the relative locations between dynamic pressure lines. Dotted yellow lines and associated text describe the main extracted features}
\label{fig:features_diagram}
\end{figure*}

Each of the decomposed EPV formulation components presents challenging tasks and requires sufficiently comprehensive representations of the game states to produce accurate estimates. We build these state representations from a wide set of low-level and fine-grained features extracted from tracking data (see Section \ref{sec:epv_markov} for the definition of tracking data). While low-level features are straightforwardly obtained from this data (i.e., players' location and speed), fine-grained features are built through either statistical models or handcrafted algorithms developed in collaboration with a group of soccer match analysts from FC Barcelona. Figure \ref{fig:features_diagram} presents a visual representation of a game situation where we can observe the available players and ball locations and a subset of features derived from that tracking data snapshot. Conceptually, we split the features into two main groups: spatial features and contextual features. Both feature types are described in Section \ref{sec:spatial_features} and Section \ref{sec:contextual_features}. The full set of features and their usage within the different models presented in this work are detailed in Appendix \ref{app:feature_set}.

\subsection{Spatial Features}\label{sec:spatial_features}
We consider spatial features those directly derived from the spatial location of the players and the ball in a given time range. These can be obtained for any game situation regardless of the context and comprise mainly physical and spatial information.
Table \ref{table:spatial_features} details a set of concepts where the specific list of features presented in Appendix \ref{app:feature_set} are derived from. The main spatial features obtained from tracking data are related to the location of players from both teams, the velocity vector of each player, the ball's location, and the location of the opponent's goal at any time instance. From the player's spatial location, we produce a series of features related to the control of space and players' density along the field. The statistical models used for pitch control and pitch influence evaluation are detailed in Appendix \ref{app:pitch_control}.

\begin{table}[h!]

\begin{tabular}{lp{8cm}}
\hline\noalign{\smallskip}
Concept type & Description  \\
\noalign{\smallskip}\hline\noalign{\smallskip}
(x,y) location & Location of a player, the ball, or attempted action, normalized in the [0,1] range according to pitches' dimensions.\\
Pitch control & Probability of controlling the ball in a specific location. \\
Pitch influence & Degree of influence of a set of players in a specific location. \\
Distance between locations & Distance in meters between two locations.\\
The angle between locations & Angle in degrees between two locations.\\
Player's velocity & Player's velocity vector in the last second.\\
\noalign{\smallskip}\hline
\end{tabular}
\caption{Description of a set of spatial concepts derived from tracking data.}
\label{table:spatial_features} 
\end{table}

\subsection{Contextual Features}\label{sec:contextual_features}

To provide a more comprehensive state representation, we include a series of features derived from soccer-specific knowledge, which provides contextual information to the model. Table \ref{table:contextual_features} presents the main concepts from which multiple contextual features are derived.

\begin{table}[h!]

\begin{tabular}{lp{8cm}}
\hline\noalign{\smallskip}
Concept type & Description  \\
\noalign{\smallskip}\hline\noalign{\smallskip}
Possession & Possessions start and end times are identified to segment each match in episodes or sequences of actions. \\
Dynamic pressure lines & Relative positioning of players according to the team's current formation or the opponents. These formations change dynamically and are calculated in time ranges of a few seconds.\\
Outplayed players & Number of players that are surpassed after an action is attempted. \\
Interceptability & Features related to the likelihood of intercepting the ball. \\
Baseline event-based models & Models built on top of event-data, which are used as a baseline to enrich the learning of tracking data-based models.\\

\noalign{\smallskip}\hline
\end{tabular}
\caption{Description of a set of contextual concepts derived from tracking data.}
\label{table:contextual_features} 
\end{table}

The concept of dynamic pressure lines refers to players being aligned with their teammates within different alignment groups. For example, a typical conceptualization of pressure lines in soccer would be the groups formed by the defenders, the midfielders, and the attackers, which tend to be aligned to keep a consistent formation. The details on the calculation of dynamic pressure lines are presented in Appendix \ref{app:dynamic_pressure_lines}. By identifying the pressure lines, we can obtain every player's opponent-relative location, which provides high-level information about players' expected behavior. For example, when a player controls the ball and is behind the opponent's first pressure line, we would expect a different pressure behavior and turnover risk than when the ball is close to the third pressure line and the goal. Also, the football experts that accompanied this study considered passes that break pressure lines to significantly impact the increase of the goal expectation at the end of the possession. \\ 


From the concept of outplayed players, we can derive features such as the number of opponent players to overcome after a given pass is attempted or the number of teammates in front of or behind the ball, among many similar derivatives. In combination with the opponent's formation block location, we can obtain information about whether the pass is headed towards the inside or outside of the formation block and how many players are to be surpassed. Intuitively, a pass that outplays several players and that is headed towards the inside of the opponent block is more likely to produce an increase of the EPV, than a pass back directed outside the opponent's block that adds two more opponent players in front of the ball. On the other hand, the interceptability concept is expected to play an essential role in capturing opponents' spatial influence near a shooting option, allowing us to produce a more detailed expected goals model. Mainly, we derive features related to the number of players pressing the shooter closely and the number of players in the triangle formed between the shooter and the posts.\\

The described spatial and contextual features represent the main building blocks for deriving the set of features used for each implemented model. In Section \ref{sec:inference}, we describe in great detail the characteristics of these models.

\section{Separated component inference}\label{sec:inference}


In this section we describe in detail the approaches followed for estimating each of the components described in Equation \ref{eq:structured_epv}, \ref{eq:pass_expectation} and \ref{eq:ball_drive_expectation}.
In general, we use function approximation methods to learn models for these components from spatiotemporal data. Specifically, we want to approximate some function $f^{*}$ that maps a set of features $x$, to an outcome $y$, such that $y=f^{*}(x)$. To do this, we will find the mapping $y=f(x;\theta)$ to learn the values of a set of parameters $\theta$ that result in an approximation to $f^{*}$.\\  

Customized convolutional neural network architectures are used for estimating probability surfaces for the components involving passes, such as pass success probability, the expected possession value of passes, and the field-wide pass selection surface. Standard shallow neural networks are used to estimate ball drive probability, expected possession value from ball drives and shots, and the action selection probability components. This section describes the selection of features $x$, observed value $y$, and model parameters $\theta$ for each component.


\subsection{Estimating pass impact at every location on the field}
\label{sec:pass_models_inference}

One of the most significant challenges when modeling passes in soccer is that, in practice, passes can go anywhere on the field. 
Previous attempts on quantifying pass success probability and expected value from passes in both soccer and basketball assume that the passing options a given player has been limited to the number of teammates on the field, and centered at their location at the time of the pass \citep{power2017not,cervone2016multiresolution,hubavcek2018deep}. However, in order to accurately estimate the impact of passes in soccer (a key element for estimating the future pathways of a possession), we need to be able to make sense of the spatial and contextual information that influences the selection, accuracy, and potential risk and reward of passing to any other location on the field. We propose using fully convolutional neural network architectures designed to exploit spatiotemporal information at different scales. We extend it and adapt it to the three related passing action models we require to learn: pass success probability, pass selection probability and pass expected value. While these three problems necessitate from different design considerations, we structure the proposed architectures in three main conceptual blocks: a \emph{feature extraction block}, a \emph{surface prediction block}, and a \emph{loss computation block}. The proposed models for these three problems also share the following common principles in its design: a layered structure of input data, the use of fully convolutional neural networks for extracting local and global features and learning a surface mapping from single-pixel correspondence. We first detail the common aspects of these architectures and then present the specific approach for each of the mentioned problems.

\paragraph{Layers of low-level and field-wide input data}
To successfully estimate a full prediction surface, we need to make sense of the information at every single pixel. Let the set of locations $L$, presented in section \ref{sec:epv_markov}, be a discrete matrix of locations on a soccer field of width $w$ and height $h$, we can construct a layered representation of the game state $Y(T_t)$, 
consisting on a set of slices of location-wise data of size $w\times h$. By doing this, we define a series of layers derived from the data snapshot $T_t$ that represent both spatial and contextual low-level information for each problem. This layered structure provides a flexible approach to include all kinds of information available or extractable from the spatiotemporal data, which is considered relevant for the specific problem being addressed. 

\paragraph{Feature extractor block}
The feature extractor block is fundamentally composed of fully convolutional neural networks for all three cases, based on the SoccerMap architecture \citep{fernandez2020soccermap}. Using fully convolutional neural networks, we leverage the combination of layers at different resolutions, allowing us to capture relevant information at both local and global levels, producing location-wise predictions that are spatially aware. Following this approach, we can produce a full prediction surface directly instead of a single prediction on the event's destination. The parameters to be learned will vary according to the input surfaces' definition and the target outcome definition. However, the neural network architecture itself remains the same across all the modeled problems. This allows us to quickly adapt the architecture to specific problems while keeping the learning principles intact. A detailed description of the SoccerMap architecture is presented in Appendix \ref{app:soccernet}.

\paragraph{Learning from single-pixel correspondance}
Usually, approaches that use fully convolutional neural networks have the ground-truth data for the full output surface. In more challenging cases, only a single classification label is available, and a weakly supervised learning approach is carried out to learn this mapping \citep{pathak2015constrained}. However, in soccer events, only a single pixel ground-truth information is available: for example, the destination location of a successful pass. This makes our problem highly challenging, given that there is only one single-location correspondence between input data and ground-truth. At the same time, we aim to estimate a full probability surface. Despite this extreme set-up, we show that we can successfully learn full probability surfaces for all the pass-related models. We do so by selecting a single pixel from the predicted output matrix, during training, according to the known destination location of observed passes, and back-propagating the loss at a single-pixel level.\\

In the following sections, we describe the design characteristics for the feature extraction, surface prediction, and loss computation blocks for the three pass-related problems: pass success probability, pass selection probability, and expected value from passes. By joining these models' output, we will obtain a single action-value estimation (EPV) for passing actions, expressed by $\E[G | A=\rho, T_t]$. The detailed list of features used for each model is described in Appendix \ref{app:feature_set}.

\subsubsection{Pass success probability}\label{sec:pass_probability}
\begin{figure*}
  \includegraphics[width=1\textwidth]{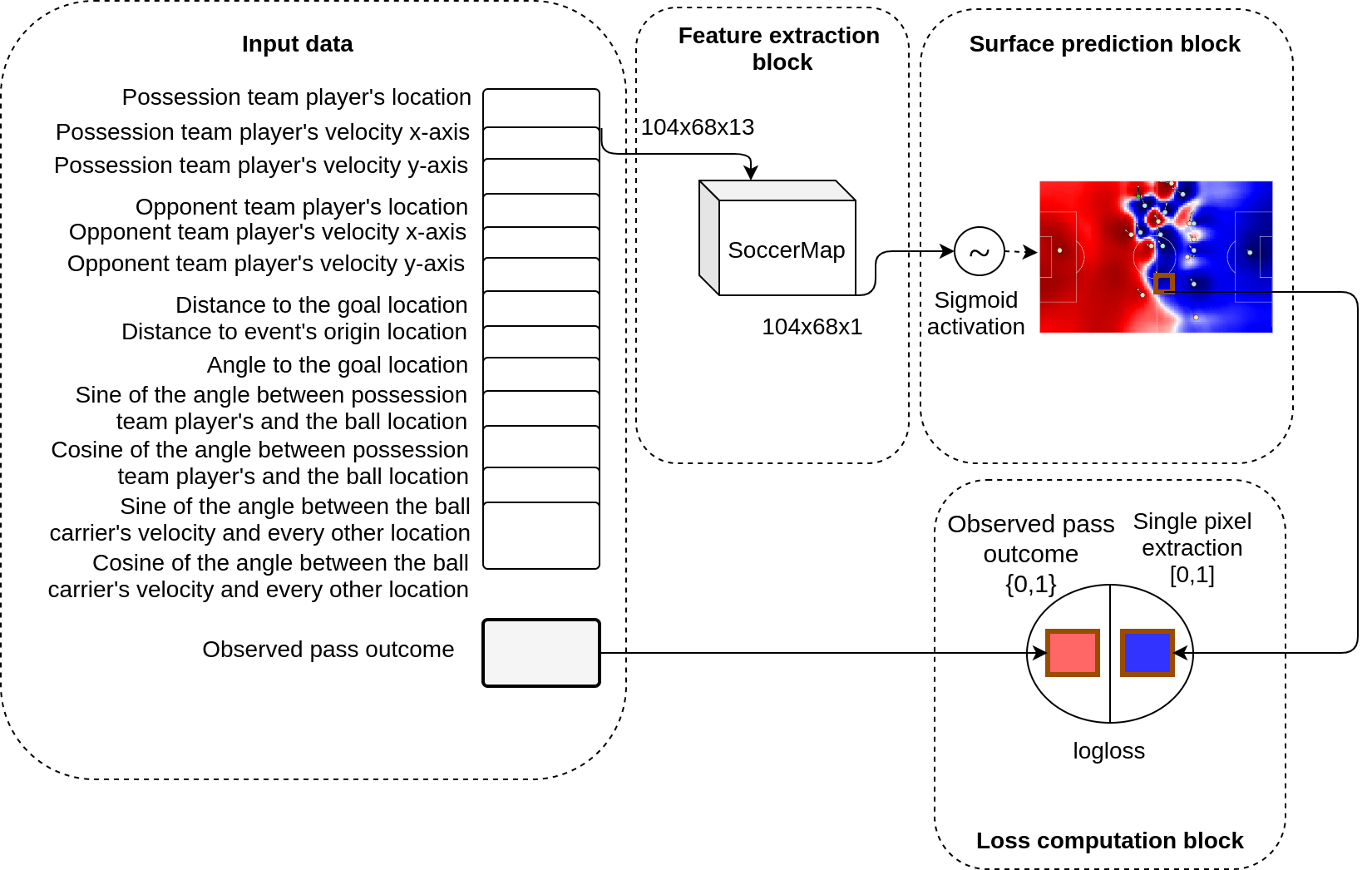}
\caption{Representation of the neural network architecture for the pass probability surface estimation, for a coarsened representation of size 104$\times$68. Thirteen layers of spatial features are fed to a SoccerMap feature extraction block, which outputs a 104$\times$68$\times$1 prediction surface. A sigmoid activation function is applied to each output, producing a pass probability surface. The output at the destination location of an observed pass is extracted, and the log loss between this output and the observed outcome of the pass is back-propagated to learn the network parameters.}
\label{fig:pass_probability_diagram}
\end{figure*}

From any given game situation where a player controls the ball, we desire to estimate the success probability of a pass attempted towards any other of the potential destination locations, expressed by $\PP(A=\rho, D_t | T_t)$. Figure \ref{fig:pass_probability_diagram} presents the designed architecture for this problem. The input data at time $t$ is conformed by 13 layers of spatiotemporal information obtained from the tracking data snapshot $T_t$ consisting mainly of information regarding the location, velocity, distance, and angles between the both team's players and the goal. The feature extraction block is composed strictly by the SoccerMap architecture, where representative features are learned. This block's output consists of a $104\times68\times1$ pass probability predictions, one for each possible destination location in the coarsened field representation. In the surface prediction block a sigmoid activation function $\sigma$ is applied to each prediction input to produce a matrix of pass probability estimations in the [0,1] continuous range, where $\sigma(x) = \frac{e^x}{e^x+1}$. Finally, at the loss computation block, we select the probability output at the known destination location of observed passes and compute the negative log loss, defined in Equation  \ref{eq:negative_logloss}, between the predicted ($\hat{y}$) and observed pass outcome ($y$).\\

\begin{equation}\label{eq:negative_logloss}
\mathcal{L}(\hat{y},y) = - (y \cdot \log(\hat{y}) + (1-y) \cdot \log(1-\hat{y}))
\end{equation}

Note that we are learning all the network parameters $\theta$ needed to produce a full surface prediction by the back-propagation of the loss value between the predicted value at that location and the observed outcome of pass success at a single location. We show in Section \ref{sec:results} that this learning set is sufficient to obtain remarkable results.

\subsubsection{Expected possession value from passes}\label{sec:pass_epv}

Once we have a pass success probability model, we are halfway to obtaining an estimation for $\E[G|A=\rho, D_t, T_t]$, as expressed in Equation \ref{eq:pass_expectation}. The remaining two components, 
$\E[G|A=\rho, O_p=1, Dt, Tt]$ and $\E[G|A=\rho, O_p=0, Dt,Tt]$, correspond to the expected value of successful and unsuccessful passes, respectively. We learn a model for each expression separately; however, we use an equivalent architecture for both cases. The main difference is that one model must be learned with successful passes and the other with missed passes exclusively to obtain full surface predictions for both cases.\\

The input data matrix consists of 16 different layers with equivalent location, velocity, distance, and angular information to those selected for the pass success probability model. Additionally, we append a series of layers corresponding to contextual features related to outplayed players' concepts and dynamic pressure lines. Finally, we add a layer with the pass probability surface, considering that this can provide valuable information to estimate the expected value of passes. This surface is calculated by using a pre-trained version of a model for the architecture presented in Section \ref{sec:pass_probability}.\\

The input data is fed to a SoccerMap feature extraction block to obtain a single prediction surface. In this case, we must observe that the expected value of $G$ should reside within the $[-1,1]$ range, as described in Section \ref{sec:epv_markov}. To do so, in the surface prediction block, we apply a sigmoid activation function to the SoccerMap predicted surface obtaining an output within $[0,1]$. We then apply a linear transformation, so the final prediction surface consists of values in the $[-1,1]$ range. Notably, our modeling approach does not assume that a successful pass must necessarily produce a positive reward or that missed passes must produce a negative reward.\\


The loss computation block computes the mean squared error between the predicted values and the reward assigned to each pass, defined in Equation \ref{eq:mse}. The model design is independent of the reward choice for passes. In this work, we choose a long-term reward associated with the observed outcome of the possession, detailed in Section \ref{sec:estimands}.

\begin{equation}\label{eq:mse}
\text{MSE}(\hat{y},y) =  \frac{1}{N} \sum_i^N(y_i-\hat{y}_i)^2
\end{equation}


\subsubsection{Pass selection probability}

Until now, we have models for estimating both the probability and expected value surfaces for both successful and missed passes.
In order to produce a single-valued estimation of the expected value of the possession given a pass is selected, we model the pass selection probability $\PP(A=\rho, D_t | T_t)$ as defined in Equation \ref{eq:structured_epv}. The values of a pass selection probability surface must necessarily add up to 1, and will serve as a weighting matrix for obtaining the single estimate.\\

Both the input and feature extraction blocks of this architecture are equivalent to those designed for the pass success probability model (see Section \ref{sec:pass_probability}). However, we use the softmax activation function presented in Equation \ref{eq:softmax} for the surface prediction block, instead of a sigmoid activation function.
We then extract the predicted value at a given pass destination location and compute the log loss between that predicted value and 1, since only observed passes are used. With the different models presented in Section \ref{sec:pass_models_inference}, we can now provide a single estimate of the expected value given a pass action is selected, $\E[G|A=\rho, T_t]$.

\begin{equation}\label{eq:softmax}
\textup{softmax}(v)_i = \frac{e^{v_i}}{\sum_{j=1}^{K}e^{v_i}} \text{for } i=0,\ldots, K
\end{equation}


\subsection{Estimating ball drive probability}\label{sec:drive_prob}

We will focus now on the components needed for estimating the expected value of ball drive actions. In this work's scope, a ball drive refers to a one-second action where a player keeps the ball in its possession. Moreover, when a player attempts a ball drive, we assume the player will maintain its velocity, so the event's destination location would be the player's expected location in the next second. While keeping the ball, the player might sustain the ball-possession or lose the ball (either because of bad control, an opponent interception, or by driving the ball out of the field, among others). The probability of keeping control of the ball with these conditions is modeled by the expression $\PP(O_{\delta}=1 | A=\delta, T_t)$.\\

We use a standard shallow neural network architecture to learn a model for this probability, consisting of two fully-connected layers, each one followed by a layer of ReLu activation functions, with a single-neuron output preceded by a sigmoid activation function. We provide a state representation for observed ball drive actions that are composed of a set of spatial and contextual features, detailed in Appendix \ref{app:feature_set}. Among the spatial features, the level of pressure a player in possession of the ball receives from an opponent player is considered to be a critical piece of information to estimate whether the possession is maintained or lost. We model pressure through two additional features: the opponent's team density at the player's location and the overall team pitch control at that same location. Another factor that is considered to influence the ball drive probability is the player's contextual-relative location at the moment of the action. We include two features to provide this contextual information: the closest opponent's vertical pressure line and the closest possession team's vertical pressure line to the player. These two variables are expected to serve as a proxy for the opponent's pressing behavior and the player's relative risk of losing the ball. 
By adding features related to the spatial pressure, we can get a better insight into how pressed that player is within that context and then have better information to decide the probability of keeping the ball. We train this model by optimizing the loss between the estimated probability and observed ball drive actions that are labeled as successful or missed, depending on whether the ball carrier's team can keep the ball's possession during after the ball drive is attempted.

\subsection{Estimating ball drive expectation}\label{sec:balldrive_expectation}

Finally, once we have an estimate of the ball drive probability, we still need to obtain an estimate of the expected value of ball drives, in order to model the expression $\E[G|A=\delta,T_t]$, presented in Equation \ref{eq:ball_drive_expectation}. While using a different architecture for feature extraction, we will model both $\E[G|A=\delta,O_\delta=1,T_t]$ and $\E[A=\delta,O_\delta=0,T_t]$, following an analogous approach of that used in Section \ref{sec:pass_epv}.\\

Conceptually, by keeping the ball, player's might choose to continue a progressive run or dribble to gain a better spatial advantage. However, they might also wait until a teammate moves and opens up a passing line of lower risk or higher quality. By learning a model for the expression $\E[G| A=\delta, T_t]$ we aim to capture the impact on the expected possession value of these possible situations, all encapsulated within the ball drive event. 
We use the same input data set and feature extractor architecture used in Section \ref{sec:drive_prob}, with the addition of the ball drive probability estimation for each example. Similarly to the loss surface prediction block of the expected value of passes (see Section \ref{sec:pass_epv}), we apply a sigmoid activation function to obtain a prediction in the $[0,1]$ range, and then apply a linear transformation to produce a prediction value in the $[-1,1]$ range. The loss computation block computes the mean squared loss between the observed reward value assigned to the action and the model output.


\subsection{Expected goals model}\label{sec:expected_goals_model}

Once we have a model for the expected values of passes and ball drives, we only need to model the expected value of shots to obtain a full value state-value estimation for the action set $A$. We want to model the expectation of scoring a goal at time $t$ given that a shot is attempted, defined as $\E[G|A=\varsigma]$. This expression is typically referred to as \emph{expected goals} (xG) and is arguably one of the most popular metrics in soccer analytics \citep{eggels2016expected}. While existing approaches make use exclusively of features derived from the observed shot location, here we include both spatial and contextual information related to the other 22 players' and the ball's locations to account for the nuances of shooting situations.\\

Intuitively, we can identify several spatial factors that influence the likelihood of scoring from shots, such as the level of defensive pressure imposed on the ball carrier, the interceptability of the shot by the nearby opponents, or the goalkeeper's location. Specifically, we add the number of opponents that are closer than 3 meters to the ball-carrier to quantify the level of immediate pressure on the player. Additionally, we account for the interceptability of the shot (blockage count) by calculating the number of opponent players in the triangle formed by the ball-carrier location and the two posts. 
We include three additional features derived from the location of the goalkeeper. The goalkeeper's location can be considered an important factor influencing the scoring probability, particularly since he has the considerable advantage of being the only player that can stop the ball with his hands. 
In addition to this spatial information, we add a contextual feature consisting of a boolean flag indicating whether the shot is taken with the foot or the head, the latter being considered more difficult. Additionally, we add a prior estimation of expected goal as an input feature to this spatial and contextual information, produced through the baseline expected goals model described in Appendix \ref{app:xg}. The full set of features is detailed in Appendix \ref{app:feature_set}. \\

Having this feature set, we use a standard neural network architecture with the same characteristics as the one used for estimating the ball drive probability, explained in Section \ref{sec:drive_prob}, and we optimize the mean squared error between the predicted outcome and the observed reward for shot actions. The long-term reward chosen for this work is detailed in Section \ref{sec:estimands}.

\subsection{Action selection probability}
Finally, to obtain a single-valued estimation of EPV we weigh the expected value of each possible action with the respective probability of taking that action in a given state, as expressed in Equation \ref{eq:structured_epv}. Specifically, we estimate the action selection probability $\PP(A | T_t)$, where $A$ is the discrete set of actions described in Section \ref{sec:epv_markov}. We construct a feature set composed of both spatial and contextual features. Spatial features such as the ball location and the distance and angle to the goal provide information about the ball carrier's relative location in a given time instance. Additionally, we add spatial information related to the possession and team's pitch control and the degree of spatial influence of the opponent team near the ball. On the other hand, the location of both team's dynamic lines relative to the ball location provides the contextual information to the state representation. We also include the baseline estimation of expected goals at that given time, which is expected to influence the action selection decision, especially regarding shot selection. The full set of features is described in Appendix \ref{app:feature_set}. We use a shallow neural network architecture, analogous to those described in Section \ref{sec:drive_prob} and Section \ref{sec:balldrive_expectation}. This final layer of the feature extractor part of the network has size 3, to which a softmax activation function is applied to obtain the probabilities of each action. We model the observed outcome as a one-hot encoded vector of size $3$, indicating the action type observed in the data, and optimize the categorical cross-entropy between this vector and the predicted probabilities, which is equivalent to the log loss.

\section{Experimental setup}
\subsection{Datasets}\label{sec:datasets}

We build different datasets for each of the presented models based on optical tracking data and event-data from 633 English Premier League matches from the 2013/2014 and 2014/2015 season, provided by \emph{STATS LLC}. This tracking data source consists of every player's location and the ball at a 10\emph{Hz} sampling rate, obtained through semi-automated player and ball tracking performed on match videos. On the other hand, event-data consists of human-labeled on-ball actions observed during the match, including the time and location of both the origin and destination of the action, the player who takes action, and the outcome of the event. Following our model design, we will focus exclusively on the pass, ball drive, and shot events.
Table \ref{table:events} presents the total count for each of these events according to the dataset split presented below in Section \ref{sec:model_setting}.
The definition of success varies from one event to another: a pass is successful if a player of the same team receives it, a ball drive is successful if the team does not lose the possession after the action occurs, and a shot is labeled as successful if a goal is scored from that shot. Given this data, we can extract the tracking data snapshot, defined in Section \ref{sec:epv_markov}, for every instance where any of these events are observed. From there, we can build the input feature sets defined for each of the presented models. For the detailed list of features used, see Appendix \ref{app:feature_set}.

\begin{table}[h!]

\resizebox{\textwidth}{!}{\begin{tabular}{llllll} 
\hline\noalign{\smallskip}
Data Type    & \# Total & \# Training & \# Validation & \# Test & \% Success  \\ 
\noalign{\smallskip}\hline\noalign{\smallskip}
Match      & 633      & 379         & 127           & 127     & -           \\
Pass       & 480,670   & 288,619      & 96,500         & 95,551   & 79.64        \\
Ball drive  & 413,123   & 284,759      & 82,271         & 82,093   & 90.60        \\
Shot       & 13,735    & 8,240        & 2,800          & 2,695    & 8.54         \\
\noalign{\smallskip}\hline
\end{tabular}}
\caption{Total count of events included within the tracking data of 633 English Premier League matches from the 2013/2014 and 2014/2015 season.}
\label{table:events}
\end{table}

\subsection{Defining the estimands}\label{sec:estimands}

Each of the components of the EPV structured model has different estimands or outcomes. For both the pass success and ball drive success probability models, we define a binomially distributed outcome, according to the definition of success provided in  \ref{sec:datasets}. These outcomes correspond to the short-term observed success of the actions. For the pass selection probability, we define the outcome as a binomially distributed random variable as well, where a value of 1 is given for every observed pass in its corresponding destination location.  We define the action selection model's estimand as a multinomially distributed random variable that can take one of three possible values, according to whether the selected action corresponds to a pass, a ball drive, or a shot.\\

For the EPV estimations of passes, ball drives, and shot actions, respectively, we define the estimand is a long-term reward, corresponding to the outcome of the possession where that event occurs. For doing this, we first need to define when possession ends. There is a low frequency of goals in matches (2.8 goals on average in our dataset) compared to the number of observed actions (1,433 on average). Given this, the definition of the time extent of possession is expected to influence the balance between individual actions' short-term value and the long-term expected outcome after that action is taken. The standard approach for setting a possession's ending time is when the ball changes control from one team to another. However, here we define a possession end as the time when the next goal occurs. By doing this, we allow the ball to either go out of the field or change control between teams an undefined number of times until the next goal is observed. Once a goal is observed, all the actions between the goal and the previous one are assigned an outcome of $1$ if the action is taken by the scoring team or $-1$ otherwise. Following this, each action gets assigned as an outcome a long-term reward (i.e., the next goal observed).\\

However, this approach is expected to introduce noise, especially for actions that are largely apart in time from an observed goal. Let $\epsilon$  be a constant representing the time between each action and the next goal, in seconds. We can choose a value for $\epsilon$ that represents a long-term reward-vanishing threshold so that all the actions observed more than $\epsilon$ time from the observed goal received a reward of $0$. For this work, we choose $\epsilon=15s$, which corresponds to the average duration of standard soccer possessions in the available matches. Note this is equivalent to assuming that the current state of a possession only has $\epsilon$ seconds impact.

\subsection{Model setting}\label{sec:model_setting}
We randomly sample the available matches and split them into training (379), validation (127), and test sets (127). From each of these matches, we obtain the observed on-ball actions and the tracking data snapshots to construct the set of input features corresponding to each model, detailed in Appendix \ref{app:feature_set}. The events are randomly shuffled in the training dataset to avoid bias from the correlation between events that occur close in time.
We use the validation set for model selection and leave the test set as a hold-out dataset for testing purposes. We train the models using the adaptive moment estimation algorithm \citep{kingma2014adam}, and set the $\beta_1$ and $\beta_2$ parameters to $0.9$ and $0.999$ respectively. For all the models we perform a grid search on the learning rate ($\{1\mathrm{e}{-3}, 1\mathrm{e}{-4}, 1\mathrm{e}{-5}, 1\mathrm{e}{-6}\}$), and batch size parameters ($\{16,32\}$). We use early stopping with a delta of $1\mathrm{e}{-3}$ for the pass success probability, ball drive success probability, and action selection probability models, and $1\mathrm{e}{-5}$ for the rest of the models.

\subsection{Model calibration}
We include an after-training calibration procedure within the processing pipeline for the pass success probability and pass selection probability models, which presented slight calibration imbalances on the validation set. We use the temperature scaling calibration method for both models, a useful approach for calibrating neural networks \citep{guo2017calibration}. Temperature scaling consists of dividing the vector of logits passed to a softmax function by a constant \emph{temperature} value $T_p$. This product modifies the scale of the probability vector produced by the softmax function. However, it preserves each element's ranking, impacting only the distribution of probabilities and leaving the classification prediction unmodified. We apply these post-calibration procedures exclusively on the validation set.

\subsection{Evaluation Metrics}\label{sec:metrics}
For the pass success probability, keep ball success probability, pass selection probability, and action selection models, we use the cross-entropy loss. Let $M$ be the number of classes, $N$ the number of examples, $y_{ij}$ the estimated outcome, and $\hat{y}_{ij}$ the expected outcome, we define the cross-entropy loss function as in Equation \ref{eq:cross_entropy}. For the first three models, where the outcome is binary, we set $M=2$. We can directly observe that for this set-up, the cross-entropy is equivalent to the negative log-loss defined in Equation \ref{eq:negative_logloss}. For the action selection model, we set $M=3$. For the rest of the models, corresponding to EPV estimations, we can observe the outcome takes continuous values in the $[-1,1]$ range. For these cases, we use the mean squared error (MSE) as a loss function, defined in Equation \ref{eq:mse}, by first normalizing both the estimated and observed outcomes into the $[0,1]$ range.

\begin{equation}\label{eq:cross_entropy}
\text{CE}(\hat{y},y) =  - \sum_j^M\sum_i^N(y_{ij} \cdot \log(\hat{y}_{ij}))
\end{equation}

We are interested in obtaining calibrated predictions for all of the models, as well as for the joint EPV estimation. Having the models calibrated allows us to perform a fine-grained interpretation of the variations of EPV within subsets of actions, as shown in Section \ref{sec:applications}. We validate the model's calibration using a variation of the expected calibration error (ECE) presented in \cite{guo2017calibration}. For obtaining this metric, we distribute the predicted outcomes into $K$ bins and compute the difference between the average prediction in each bin and the average expected outcome for the examples in each bin. Equation \ref{eq:ece} presents the ECE metric, where $K$ is the number of bins, and $B_k$ corresponds to the set of examples in the $k$-th bin. Essentially, we are calculating the average difference between predicted and expected outcomes, weighted by the number of examples in each bin. In these experiments, we use quantile binning to obtain K equally-sized bins in ascending order.

\begin{equation}\label{eq:ece}
\text{ECE} = \sum_{k=1}^{K} \frac{\abs{B_k}}{N} \abs{ \bigg(\frac{1}{|B_k|} \sum_{i \in B_k}y_i\bigg) - 
\bigg(\frac{1}{|B_k|} \sum_{i \in B_k} \hat{y}_i  \bigg)  }
\end{equation}

\subsection{Results}\label{sec:results}

Table \ref{tab:results} presents the results obtained in the test set for each of the proposed models. The loss value corresponds to either the cross-entropy or the mean squared loss, as detailed in Section \ref{sec:metrics}. 
The table includes the optimal values for the batch size and learning rate parameters, the number of parameters of each model, and the number of examples per second that each model can predict.\\

We can observe that the loss value reported for the final joint model is equivalent to the losses obtained for the EPV estimations of each of the three types of action types, showing stability in the model composition. The shot EPV loss is higher than the ball drive EPV and pass EPV losses, arguably due to the considerably lower amount of observed events available in comparison with the rest, as described in Section \ref{sec:datasets}. While the number of examples per second is directly dependent on the models' complexity, we can observe that we can predict 899 examples per second in the worst case. This value is  89 times higher than the sampling rate of the available tracking data (10Hz), showing that this approach can be applied for the real-time estimation of EPV and its components.\\

\begin{figure*}
  \includegraphics[width=1\textwidth]{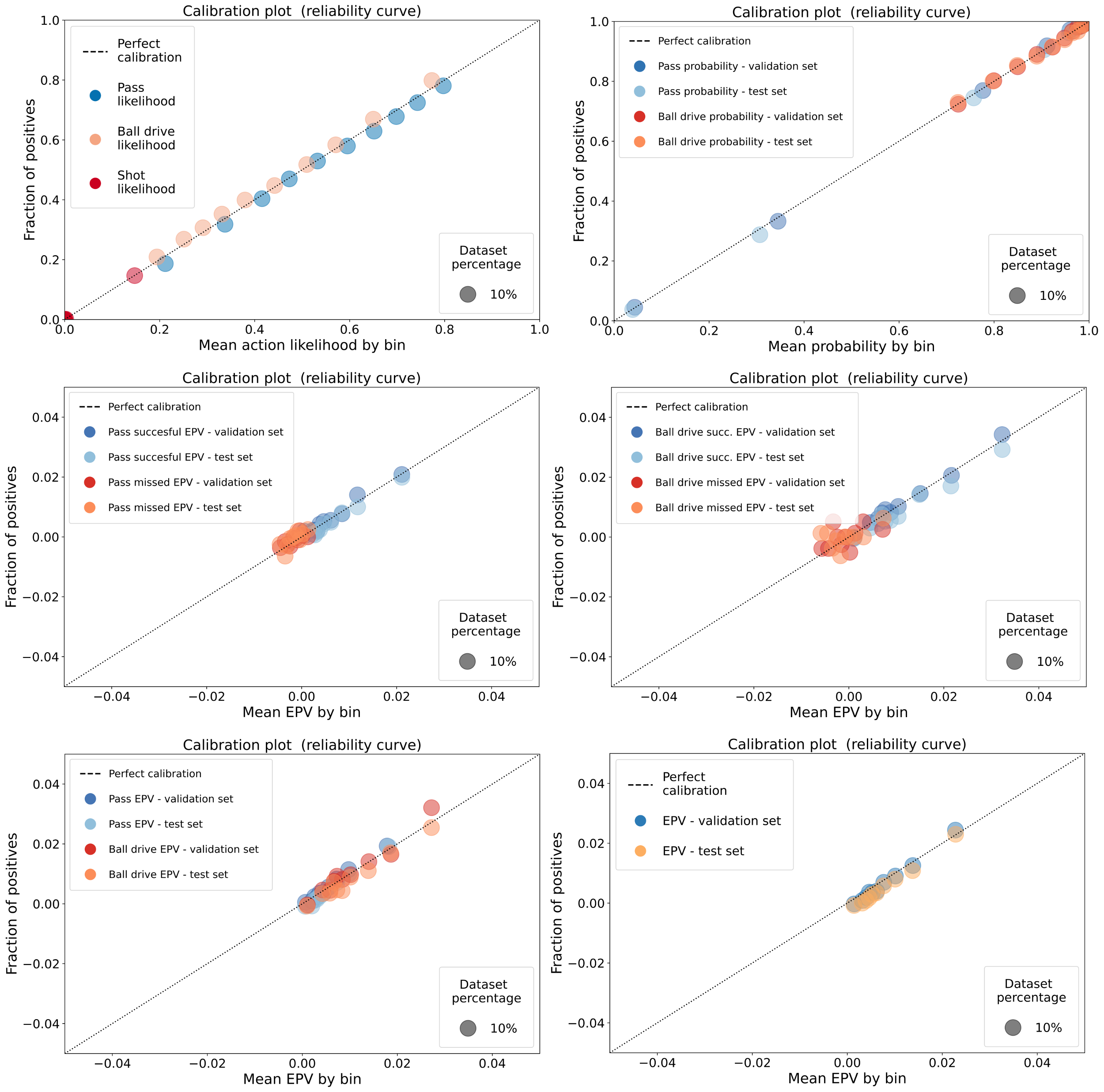}
\caption{Probability calibration plots for the action selection (top-left), pass and ball drive probability (top-right), pass (successful and missed) EPV (mid-left), ball drive (successful and missed) EPV (mid-right), pass and ball drive EPV joint estimation (bottom-left), and the joint EPV estimation (bottom-right). Values in the x-axis represent the mean value by bin, among 10 equally-sized bins. The y-axis represents the mean observed outcome by bin. The circle size represents the percentage of examples in each bin relative to the total examples for each model.}
\label{fig:calibration_all}
\end{figure*}

Regarding the models' calibration, we can observe that the ECE metrics present consistently low values along with all the models. Figure \ref{fig:calibration_all} presents a fine-grained representation of the probability calibration of each of the models. The x-axis represents the mean predicted value for a set of $K=10$ bins, while the y-axis represents the mean observed outcome among the examples within each corresponding bin. The circle size represents the percentage of examples in the bin relative to the total number of examples. In these plots, we can observe that the different models provide calibrated probability estimations along their full range of predictions, which is a critical factor for allowing a fine-grained inspection of the impact that specific actions have on the expected possession value estimation. Additionally, we can observe the different ranges of prediction values that each model produces. For example, ball drive success probabilities are distributed more often above $0.5$, while pass success probabilities cover a wide range between $0$ and $1$, showing that it is harder for a player to lose the ball while keeping possession than it is to lose the ball by attempting a pass towards another location on the field. The action selection probability distribution is heavily influenced by each action type's frequency, showing a higher frequency and broader distribution on ball drive and pass actions compared with shots. The joint EPV model's calibration plot shows that the proposed approach of estimating the different components separately and then merging them back into a single EPV estimation provides calibrated estimations. We applied post-training calibration exclusively to the pass success probability and the pass selection probability models, obtaining a temperature value of $0.82$ and $0.5$, respectively.\\

Having this, we have obtained a framework of analysis that provides accurate estimations of the long-term reward expectation of the possession, while also allowing for a fine-grained evaluation of the different components comprised in the model.\\


\begin{table}

\resizebox{\textwidth}{!}{\begin{tabular}{lllllll} 
\hline\noalign{\smallskip}
Model                 & Loss    & ECE    & \begin{tabular}[c]{@{}l@{}}Batch\\Size\end{tabular} & \begin{tabular}[c]{@{}l@{}}Learning\\Rate\end{tabular} & \# Params. & Ex. (s)  \\ 
\noalign{\smallskip}\hline\noalign{\smallskip}
Pass probability      & 0.190  & 0.0047  & 32 & $1\mathrm{e}{-4}$ &  401,259    &  942    \\
Ball drive probability & 0.2803  & 0.0051 & 32 & $1\mathrm{e}{-3}$  &  128       &  67,230  \\
Pass successful EPV              & 0.0075 & 0.0011 & 16 & $1\mathrm{e}{-6}$  &  403,659     &  899    \\
Pass missed EPV              & 0.0085 & 0.0015 & 16 & $1\mathrm{e}{-6}$  &   403,659    &     899  \\
Pass selection probability       & 5.7134   & -      & 32 & $1\mathrm{e}{-5}$  &  401,259    &   984   \\
Pass EPV * Pass selection & 0.0067  & 0.0011 & - & - & -          &     -  \\
Ball drive successful EPV         & 0.0128 & 0.0022 & 16 & $1\mathrm{e}{-4}$ &  153       &   57,441  \\
Ball drive missed EPV         & 0.0072 & 0.0025 & 16 & $1\mathrm{e}{-4}$ &     153    & 57,441  \\
Shot EPV              & 0.2421  & 0.0095  & 16 & $1\mathrm{e}{-3}$ &   231      &   72,455  \\
Action selection probability     & 0.6454  & -  & 32 & $1\mathrm{e}{-3}$ &     171    &   23,709  \\
EPV & 0.0078 & 0.0023 & - & - & - & -\\

\hline\noalign{\smallskip}

\end{tabular}}
\caption{The average loss and calibration value for each of the  components of the EPV model, as well as for the joint EPV estimation, on the corresponding test datasets. Additionally, the table presents the optimal value of the hyper-parameters, total number of parameters, and the number of predicted examples by second, for each of the models.}
\label{tab:results}
\end{table}

\section{Practical Applications}\label{sec:applications}

In this section, we present a series of novel practical applications derived from the proposed EPV framework. We show how the different components of our EPV representation can be used to obtain direct insight in specific game situations at any frame during a match. We present the value distribution of different soccer actions and the contextual features developed in this work and analyze the risk and reward comprised by these actions. Additionally, we leverage the pass EPV surfaces, and the contextual variables developed in this work to analyze different off-ball pressing scenarios for breaking Liverpool's organized buildup. Finally, we inspect the on-ball and off-ball value-added between every Manchester City player (season 14-15) and the legendary attacking midfielder David Silva, to derive an optimal team that would maximize Silva's contribution to the team.

\subsection{A real-time control room}

\begin{figure*}[h]
  \includegraphics[width=1\textwidth]{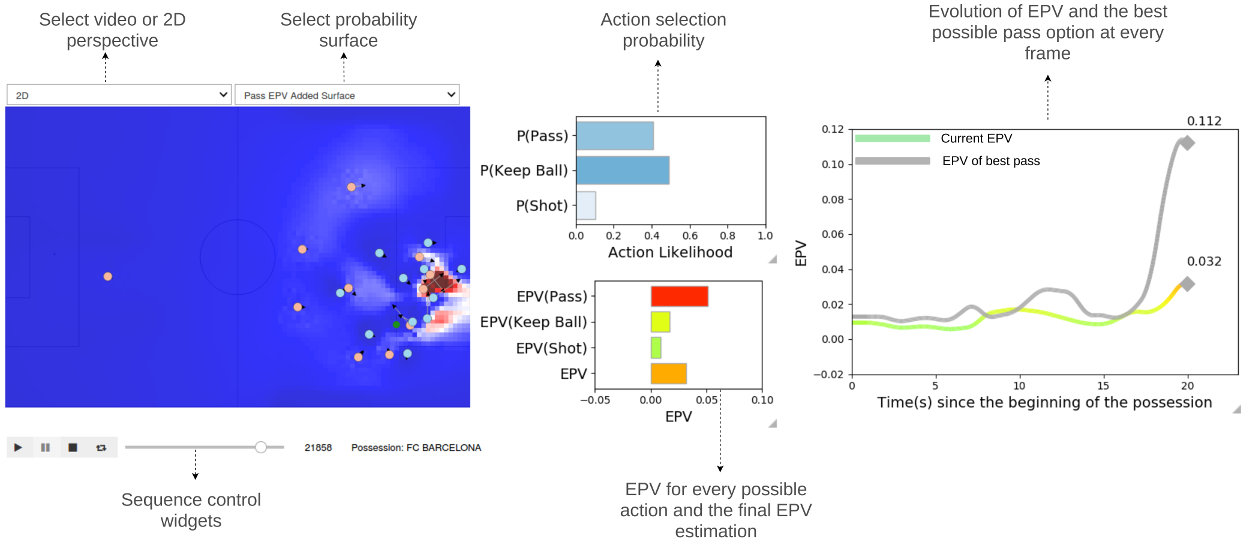}
\caption{A visual control room tool based on the EPV components. On the left, a 2D representation of the game state at a given frame during the match, with an overlay of the pass EPV added surface and selection menus to change between 2D and video perspective, and to modify the surface overlay. On the bottom-left corner, a set of video sequence control widgets. On the center, the instantaneous value of selection probability of each on-ball action, and the expected value of each action, as well as the overall EPV value. On the right, the evolution of the EPV value during the possession and the expected EPV value of the optimal passing option at every frame.}
\label{fig:app_control_room}
\end{figure*}

In most team sports, coaches make heavy use of video to analyze player performance, show players their correctly or incorrectly performed actions, and even point out other possible decisions the player may have taken in a given game situation. The presented structured modeling approach of the EPV provides the advantage of obtaining numerical estimations for a set of game-related components, allowing us to understand the impact that each of them has on the development of each possession. Based on this, we can build a control room-like tool like the one shown in Figure \ref{fig:app_control_room}, to help coaches analyze game situations and communicate effectively with players.

The control room tool presented in Figure \ref{fig:app_control_room} shows the frame-by-frame development of each of the EPV components. Coaches can observe the match's evolution in real-time and use a series of widgets to inspect into specific game situations. 
For instance, in this situation, coaches can see that passing the ball has a better overall expected value than keeping the ball or shooting. Additionally, they can visualize in which passing locations there is a higher expected value. The EPV evolution plot on the right shows that while the overall EPV is $0.032$, the best possible passing option is expected to increase this value up to $0.112$. The pass EPV added surface overlay shows that an increase of value can be expected by passing to the teammates inside the box or passing to the teammate outside the box. With this information and their knowledge on their team, coaches can decide whether to instruct the player to take immediate advantage of these kinds of passing opportunities or wait until better opportunities develop. 
Additionally, the player can gain a more visual understanding of the potential value of passing to specific locations in this situation instead of taking a shot. If the player tends shooting in these kinds of situations, the coach could show that keeping the ball or passing to an open teammate has a better goal expectancy than shooting from that location.\\

This visual approach could provide a smoother way to introduce advanced statistics into a coaching staff analysis process. Instead of evaluating actions beforehand or only delivering hard-to-digest numerical data, we provide a mechanism to enhance coaches' interpretation and player understanding of the game situations without interfering with the analysis process. 
 
\subsection{Not all value is created (or lost) equal}

There is a wide range of playing strategies that can be observed in modern professional soccer. There is no single best strategy found in successful teams from Guardiola's creative and highly attacking FC Barcelona to Mourinho's defensive and counter-attacking Inter Milan. We could argue that a critical element for selecting a playing strategy lies in managing the risk and reward balance of actions, or more specifically, which actions a team will prefer in each game situation. While professional coaches intuitively understand which actions are riskier and more valuable, there is no quantification of the actual distribution of the value of the most common actions in soccer.\\

\begin{figure*}[h]
  \includegraphics[width=1\textwidth]{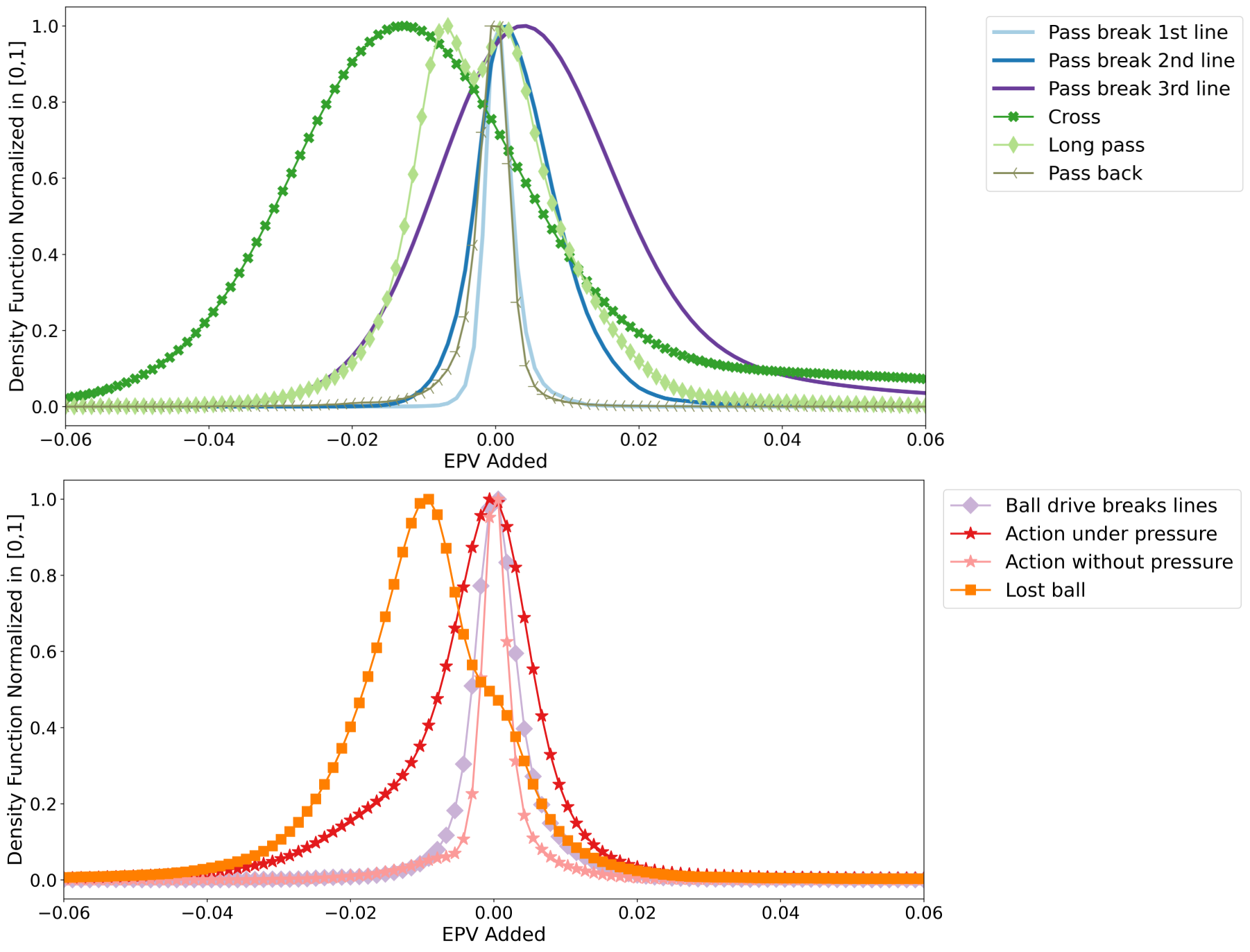}
\caption{Comparison of the probability density function of ten different actions in soccer. The density function values are normalized into the $[0,1]$ range. The normalization is obtained by dividing each density value by the maximum observed density value.}
\label{fig:app_value_creation}
\end{figure*}

From all the passes and ball drive actions described in Section \ref{sec:datasets}, and the spatial and contextual features described in Section \ref{sec:feature_extraction} we derived a series of context-specific actions to compare their value distribution. We identify passes and ball drives that break the first, second, or third line from the concept of dynamic pressure lines. 
We define an action (pass or ball drive) to be under-pressure if the player's pitch control value at the beginning of the action is below $0.4$ and without pressure otherwise. A long pass is defined as a pass action that covers a distance above 30 meters. We define a pass back as passes where the destination location is closer to the team's goal than the ball's origin location. We count with manually labeled tags indicating when a pass is a cross and when the pass is missed, from the available data. We identify lost balls as missed passes and ball drives ending in recovery by the opponent. For all of these action types, we calculate the added value of each observed action (EPV added) as the difference between the EPV at the end and the start of the action. We perform a kernel density estimation on the EPV added of each action type to obtain a probability density function. In Figure \ref{fig:app_value_creation} we compare the density between all the action types. The density function value is normalized in the $[0,1]$ range by dividing by the maximum density value in order to ease the visual comparison between the distributions.\\

From Figure \ref{fig:app_value_creation}, we can gain a deeper understanding of the value distribution of different types of actions. From passes that break lines, we can observe that the higher the line, the broader the distribution, and the higher the extreme values. While passes breaking the first line are centered around $0$ with most values ranging in $[-0.01,0.015]$, the distribution of passes breaking the third line is centered around $0.005$, and most passes fall in the interval $[-0.025,0.05]$. Similarly, ball drives that break lines present a similar distribution as passes breaking the first line.
Regarding the level of spatial pressure on actions, we can see that actions without pressure present an approximately zero-centered distribution, with most values falling in a $[-0.01,0.01]$ range. On the other hand, actions under pressure present a broader distribution and a higher density on negative values. This shows both that there is more tendency to lose the ball under pressure, hence losing value, and a higher tendency to increase the value if the pressure is overcome with successful actions. Whether crosses are a successful way for reaching the goal or not has been a long-term debate in soccer strategy. 
We can observe that crosses constitute the type of action with a higher tendency to lose significant amounts of value; however, it does provide a higher probability of high value increases in case of succeeding, compared to other actions. 
Long passes share a similar situation, where they can add a high amount of value in case of success but have a higher tendency to produce high EPV losses.
For years, soccer enthusiasts have argued about whether passing backward provides value or not. We can observe that, while the EPV added distribution of passing back is the narrowest, near half of the probability lies on the positive side of the x-axis, showing the potential value to be obtained from this type of action. Finally, losing the ball often produces a loss of value. However, in situations  such as being close to the opponent's box and with pressure on the ball carrier, losing the ball with a pass to the box might provide an increment in the expected value of the possession, given the increased chance of rebound.

\subsection{Pressing Liverpool}

A prevalent and challenging decision that coaches face in modern professional football is how to defend an organized buildup by the opponent. We consider an organized buildup as a game situation where a team has the ball behind the first pressure line. When deciding how to press, a coach needs to decide first in which zones they want to avoid the opponent receiving passes. Second, how to cluster their players in order to minimize the chances of the opponent moving forward. This section uses EPV passing components and dynamic pressure lines to analyze how to press Brendan Rodgers' Liverpool (season 14/15).\\

\begin{figure*}[h]
  \includegraphics[width=1\textwidth]{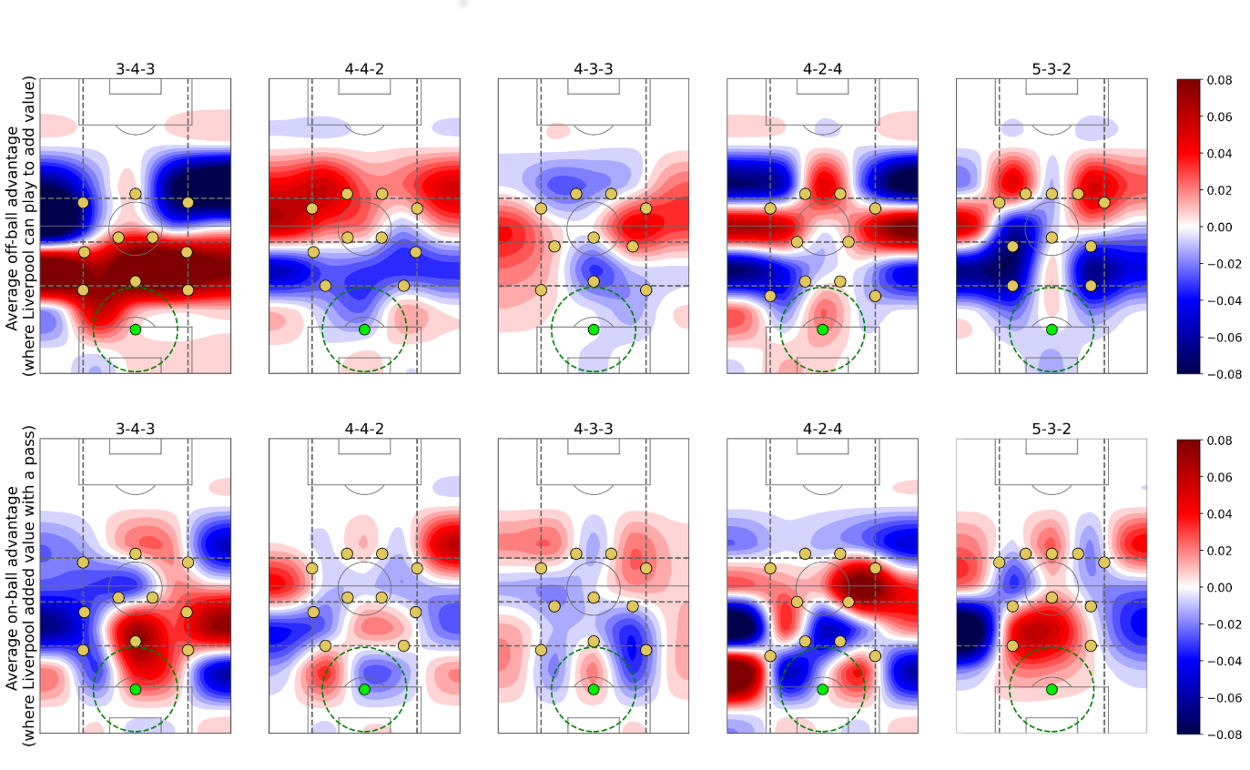}
\caption{In the first row, one distribution for every formation Liverpool's opponents used during Liverpool's organized buildups, showing the difference between the distribution of off-ball advantages and the mean distribution. The second row is analogous to the first one, presenting the on-ball EPV added distributions. The green circle represents the ball location.}
\label{fig:app_press_liverpool}
\end{figure*}

We identify the formation being used every time by counting the number of players in each pressure line. We assume there are only three pressure lines, so all formations are presented as the number of defenders followed by the number of midfielders and forwards. For every formation faced by Liverpool during buildups, we calculate both the mean off-ball and on-ball advantage in every location on the field. The on-ball advantage is calculated as the sum of the EPV added of passes with positive EPV added. On the other hand, the off-ball advantage is calculated as the sum of positive potential EPV added. We then say that a player has an off-ball advantage if he is located in a position where, in case of receiving a pass, the EPV would increase. Figure \ref{fig:app_press_liverpool} presents two heatmaps for every of the top 5 formations used against Liverpool during buildups, showing the distribution where Liverpool obtained on-ball and off-ball advantages, respectively. The heatmaps are presented as the difference with the mean heatmap in all of Liverpool's buildups during the season.

We will assume that the coach wants to avoid Liverpool playing inside its team block during buildups. We can see that when facing a 3-4-3 formation, Liverpool can create higher off-ball advantages before the second pressure line and manages to break the first line of pressure by the inside successfully. Against the 4-4-2, Liverpool has more difficulties in breaking the first line but still manages to do it successfully while also generating spaces between the defenders and midfielders, facilitating long balls to the sides. If the coaches' team does not have a good aerial game, this would be a harmful way of pressing. We can see the 4-3-3 is an ideal pressing formation for avoiding Liverpool playing inside the pressing block. This pressing style pushes the team to create spaces on the outside, before the first pressure line and after the second pressure line. In the second row, we can observe that Liverpool struggles to add value by the inside and is pushed towards the sides when passing. The 4-2-4 is the formation that avoids playing inside the block the most; however, it also allows more space on the sides of the midfielders. We can see that Liverpool can take advantage of this and create spaces and make valuable passes towards those locations. If the coach has fast wing-backs that could press receptions on long balls to the sides, this could be an adequate formation; otherwise, 4-3-3 is still preferable. Finally, the 5-3-2 provides significant advantages to Liverpool that can create spaces both by the inside above the first pressure line and behind the defenders back, while also playing towards those locations effectively.\\

This kind of information can be highly useful to a coach to decide tactical approaches for solving specific game situations. If we add the knowledge that the coach has of his players' qualities, he can make a fine-tuned design of the pressing he wants his team to develop.

\subsection{Growing around David Silva}

\begin{figure*}[h]
  \includegraphics[width=1\textwidth]{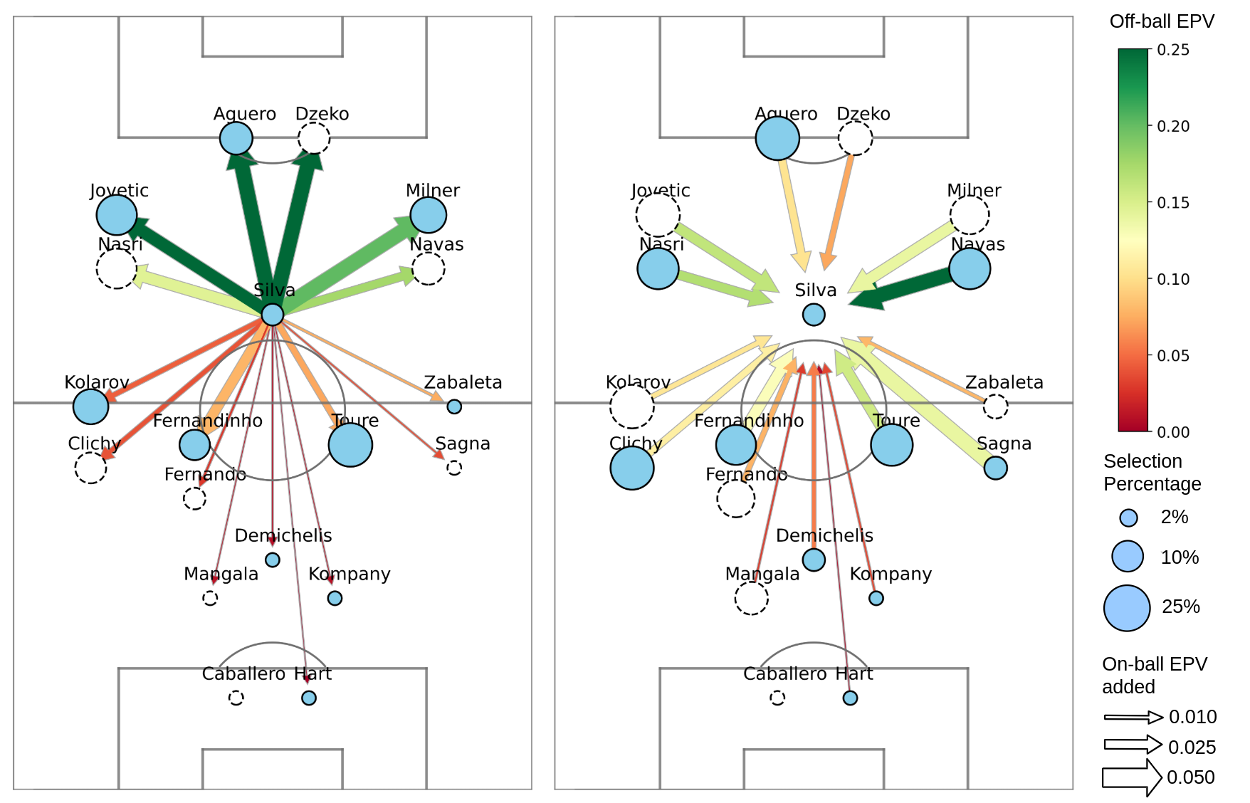}
\caption{Two passing maps representing the relationship between David Silva and each of the two players with more minutes by position in the Manchester City team during season 14/15. The figure on the left represents passes attempted by Silva, while the figure on the right represents passes received by Silva. The color of the arrow represents the average expected off-ball EPV added of the passes. The size of the circle represents the selection percentage of the destination player of the pass. Circles present a solid contour when that player is considered better for Silva than the teammate in the same position. The size of the arrow represents the mean on-ball EPV added of attempted passes. Players are placed according to their highest used position on the field.  All metrics are normalized by minutes played together and multiplied by 90 minutes.}
\label{fig:app_silva}
\end{figure*}

Most teams in the best professional soccer leagues have at least one player who is the key playmaker. Often, coaches want to ensure that the team's strategy is aligned with maximizing the performance of these key players. In this section, we leverage tracking data and the passing components of the EPV model to analyze the relationship between the well known attacking midfielder David Silva and his teammates when playing at Manchester City in season 14/15. We calculated the playing minutes each player shared with Silva and aggregated both the on-ball EPV added and expected off-ball EPV added of passes between each player pair for each match in the season. We analyze two different situations: when Silva has the ball and when any other player has the ball and Silva is on the field. We also calculate the selection percentage, defined as the percentage of time Silva chooses to pass to that player when available (and vice versa). Figure \ref{fig:app_silva} presents the sending and receiving maps involving David Silva and each of the two players with more minutes by position in the team. Every player is placed according to the most commonly used position in the league. Players represented by a circle with a solid contour have the highest sum of off-ball and on-ball EPV in each situation than the teammate assigned for the same position, presented with a dashed circle. The size of the circle represents the selection percentage of the player in each situation. We represent off-ball EPV added by the arrows' color, and on-ball EPV added of attempted passes by the arrow's size. 

We can see that both the wingers and forwards generate space for Silva and receive high added value from his passes. However, the most frequently selected player is the central midfielder Yaya Tour\'e, who also looks for Silva often and is the midfielder providing the highest value to him. Regarding the other central midfielder, Fernandinho has a better relationship with Silva in terms of received and added value than Fernando. Silva shows a high tendency to play with the wingers; however, while Milner and Jovetic can create space and receive value from Silva, Navas and Nasri find Silva more often, with higher added value. Based on this, the coach can decide whether he prefers to lineup wingers that can benefit from Silva's passes or wingers, increasing Silva's participation in the game. A similar situation is presented with the right and left-backs. Additionally, we can observe that Silva tends to be a highly preferable passing option for most players. This information allows the coach to gain a deeper understanding of the effective off-ball and on-ball value relationship that is expected from every pair of players and can be useful for designing playing strategies before a match.

\section{Discussion}

This paper presents a comprehensive approach for estimating the instantaneous expected value of ball possessions in soccer. One of the main contributions of this work is showing that by deconstructing a single expectation into a series of lower-level statistical components and then estimating each of these components separately, we can gain greater interpretation insight into how these different elements impact the final joint estimation. Also, instead of depending on a single-model approach, we can make a more specialized selection of the models, learning approach, and input information that is better suited for learning the specific problem represented by each sub-component of the EPV decomposition. The deep learning architectures presented for the different passing components produce full probability surfaces, providing rich visual information for coaches that can be used to perform fine-grained analysis of player's and team's performance. We show that we can obtain calibrated estimations for all the decomposed model components, including the single-value estimation of the expected possession value of soccer possessions. We develop a broad set of novel spatial and contextual features for the different models presented, allowing rich state representations.  Finally, we present a series of practical applications showing how this framework could be used as a support tool for coaches, allowing them to solve new upcoming questions and accelerating the problem-solving necessities that arise daily in professional soccer.\\

We consider that this work provides a relevant contribution to improving the practitioners' interpretation of the complex dynamics of professional soccer. With this approach, soccer coaches gain more convenient access to detailed statistical estimations that are unusual in their practice and find a visual approach to analyze game situations and communicate tactics to players. Additionally, on top of this framework, there is a large set of novel research that can be derived, including on-ball and off-ball player performance analysis, team performance and tactical analysis for pre-match and post-match evaluation, player profile identification for scouting, young players evolution analysis, match highlights detection, and enriched visual interpretation of game situations, among many others.

\appendix
\section{List of positional and contextual features}
\label{app:feature_set}

Table \ref{tab:features_all} describes the complete set of features used as input for each presented model. The concept type column refers to the general feature grouping described in Section \ref{sec:feature_extraction}, including a prefix indicating whether the feature is a spatial feature (SP), a contextual feature (CX), or other types (OT). Model names are presented with acronyms, including: pass success probability (PP), pass selection probability (PS), pass success and missed EPV (PE), ball drive probability (KP), ball drive success and missed EPV (KE), action selection probability (AS), and shot EPV (SE). For PP, PS, and PE models, the input features are either sparse or full matrix of $104\times68$. When the feature description indicates the value is set of every location, this input will correspond to a full matrix; otherwise, it corresponds to a sparse matrix. For the rest of the models, each feature is provided as a single variable. We refer to the team in possession of the ball as the \emph{possession team}, and its players as the \emph{possession players}, based on the definition of possession presented in Section \ref{sec:estimands}. We refer to the other team as the opponent team. All the features are normalized, assuming left to right attacking direction.

\begin{table}\caption{Description of the input features used in each of the presented models.} \label{tab:features_all}
\begin{tabular}{p{2.4cm}p{5.4cm}p{0.1cm}p{0.1cm}p{0.08cm}p{0.08cm}p{0.08cm}p{0.08cm}p{0.08cm}}
\hline\noalign{\smallskip}
Concept type                & Feature                                                                                                              & PP & PS & PE & DP & DE & AS & SE  \\
\noalign{\smallskip}\hline\noalign{\smallskip}

SP - (x,y) location         & 1 on possession players' location (x,y).                                                              & \checkmark                & \checkmark               & \checkmark        &                       &               &                   &           \\
SP - (x,y) location         & Ball location (x).                                                                                                    &                  &                 &          & \checkmark                     & \checkmark             & \checkmark                 &   \checkmark        \\
SP - (x,y) location         & 1 on opponent players' location (x,y).                                                                & \checkmark                 &  \checkmark               & \checkmark        &                       &               &                   &           \\
SP - (x,y) location         & 1 if the ball is closer to the goal than the opponent's goalkeeper.                      &                  &                 &          &                       &               &                   & \checkmark         \\
SP - Velocity               & Possession team players' speed (m/s) (x).                                                              & \checkmark                & \checkmark               & \checkmark        &                       &               &                   &           \\
SP - Velocity               & Possession team players' speed  (m/s) (y).                                                             & \checkmark                & \checkmark               & \checkmark        &                       &               &                   &           \\
SP - Velocity               & Opponent team players' speed  (m/s) (y).                                                                & \checkmark                & \checkmark               & \checkmark       &                       &               &                   &           \\
SP - Velocity               & Opponent team players' speed (m/s) (y).                                                                & \checkmark                &\checkmark               & \checkmark        &                       &               &                   &           \\
SP - Angle                  & Angle  between every location and the goal                                                       & \checkmark                &   \checkmark               & \checkmark        &                       &               &                   &           \\
SP - Angle                  & Angle between the ball and the goal.                                                    &                  &                 &          & \checkmark                     & \checkmark             & \checkmark                 & \checkmark         \\
SP - Angle                  & Sine of the angle between every location and the ball location.                                                      & \checkmark                &   \checkmark               &          &                       &               &                   &           \\
SP - Angle                  & Cosine of the angle between every location and the ball location.                                                    & \checkmark                &    \checkmark             &          &                       &               &                   &           \\
SP - Angle                  & Sine of the angle between the ball carrier velocity vector and every other location.                                                      & \checkmark                &      \checkmark           &          &                       &               &                   &           \\
SP - Angle                  & Cosine of the angle between the ball carrier velocity vector and every other location.                                                      & \checkmark                &     \checkmark            &          &                       &               &                   &           \\
SP - Distance               & Distance between every location and the goal.                                                                         & \checkmark                & \checkmark              & \checkmark        &                       &               &                   &           \\
SP - Distance               & Distance between every location and the ball.                                                                       & \checkmark                & \checkmark               & \checkmark        &                       &               &                   &           \\
SP - Distance               & Distance between the ball and the goal.                                                                      &                  &                 &          & \checkmark                     & \checkmark             & \checkmark                 & \checkmark         \\
SP - Distance               & Distance between the ball and the  goalkeeper in y-axis.                                            &                  &                 &          &                       &               &                   & \checkmark         \\
SP - Distance               & Distance between the ball and the goalkeeper.                                                      &                  &                 &          &                       &               &                   & \checkmark         \\
SP - Pitch control          & Pitch control of the possession team at the ball location                                                            &                  &                 &          & \checkmark                     & \checkmark             &\checkmark &           \\
SP - Pitch influence        & Pitch influence of the opponent team at the ball location.                                                            &                  &                 &          & \checkmark                     & \checkmark             & \checkmark                &           \\
CX - Dynamic pressure lines & Index of the closest possession team line to every location.                                                          &                  &                 & \checkmark        &                       &               &                   &           \\
CX - Dynamic pressure lines & Index of the closest possession team line to the ball location.                                                       &                  &                 &          & \checkmark                     & \checkmark             & \checkmark                 &           \\
CX - Dynamic pressure lines & Index of the closest opponent team line to every location.                                                            &                  &                 & \checkmark        &                       &               &                   &           \\
CX - Dynamic pressure lines & Index of the closest opponent team line to the ball location.                                                         &                  &                 &          & \checkmark                     & \checkmark             & \checkmark &           \\
CX - Outplayed players      & Number of possession team's players between the ball and every other location.                         &                  &                 & \checkmark        &                       &               &                   &           \\
CX - Outplayed players      & Number of opponent players between the ball and every other location.                            &                  &                 & \checkmark        &                       &               &                   &           \\
CX - Outplayed players      & Number of possession players  between the opponent's goal and every other location.                        &                  &                 & \checkmark        &                       &               &                   &           \\
CX - Outplayed players      & Number of players of the opponent team between the opponent's goal and every other location.                          &                  &                 & \checkmark        &                       &               &                   &           \\
CX - Interceptability       & Number of opponent players inside the triangle formed between the ball location and the posts of the opponent's goal. &                  &                 &          &                       &               &                   & \checkmark         \\
CX - Interceptability       & Number of opponent players located less than 3 meters away from the ball location.                                    &                  &                 &          &                       &               &                   & \checkmark         \\
OT - Type                   & 1 of action is attempted with the head.                                               &                  &                 &          &                       &               &                   & \checkmark         \\
OT -~ Event-based xG        & Expected goals based on the action location and the angle to the goal.                       &                  &                 &          &                       &               & \checkmark                 & \checkmark        
\\
OT -~ Probability       & Pass probability surface. &                  &                 &     \checkmark     &                       &               &                 &       
\\
OT -~ Probability       & Ball drive probability. &                  &                 &        &                       &              \checkmark   &                 &       
\\
\noalign{\smallskip}\hline

\end{tabular}
\end{table}

\section{Pitch control and influence model}\label{app:pitch_control}

The concepts of pitch influence and pitch control are adapted from a recent statistical approach based on modeling players' reachability surface through normal distributions \citep{fernandez2018wide}. The pitch influence $I$, shown below at expression \ref{eq:pitch_influence_random_variable} is a normally distributed random variable whose mean vector and covariance matrix are adjusted to account for players' velocity and ball location. Let $p_i$ be player's $i$ location in 1 second and let $f_i(p,t)$ be the value of the probability density function of $I$ related to player $i$ at location $p$ and time $t$, we obtain the player's influence value at location $I_i(p,t)$ following expression \ref{eq:pitch_influence}.

\begin{equation}\label{eq:pitch_influence_random_variable}
I \sim \mathcal{N}(\mu,\,\Sigma)\
\end{equation}

\begin{equation}\label{eq:pitch_influence}
I_i(p,t) = \dfrac{f_i(p,t)}{f_i(p_i,t))}.
\end{equation}

\begin{figure*}[h!]
  \includegraphics[width=1\textwidth]{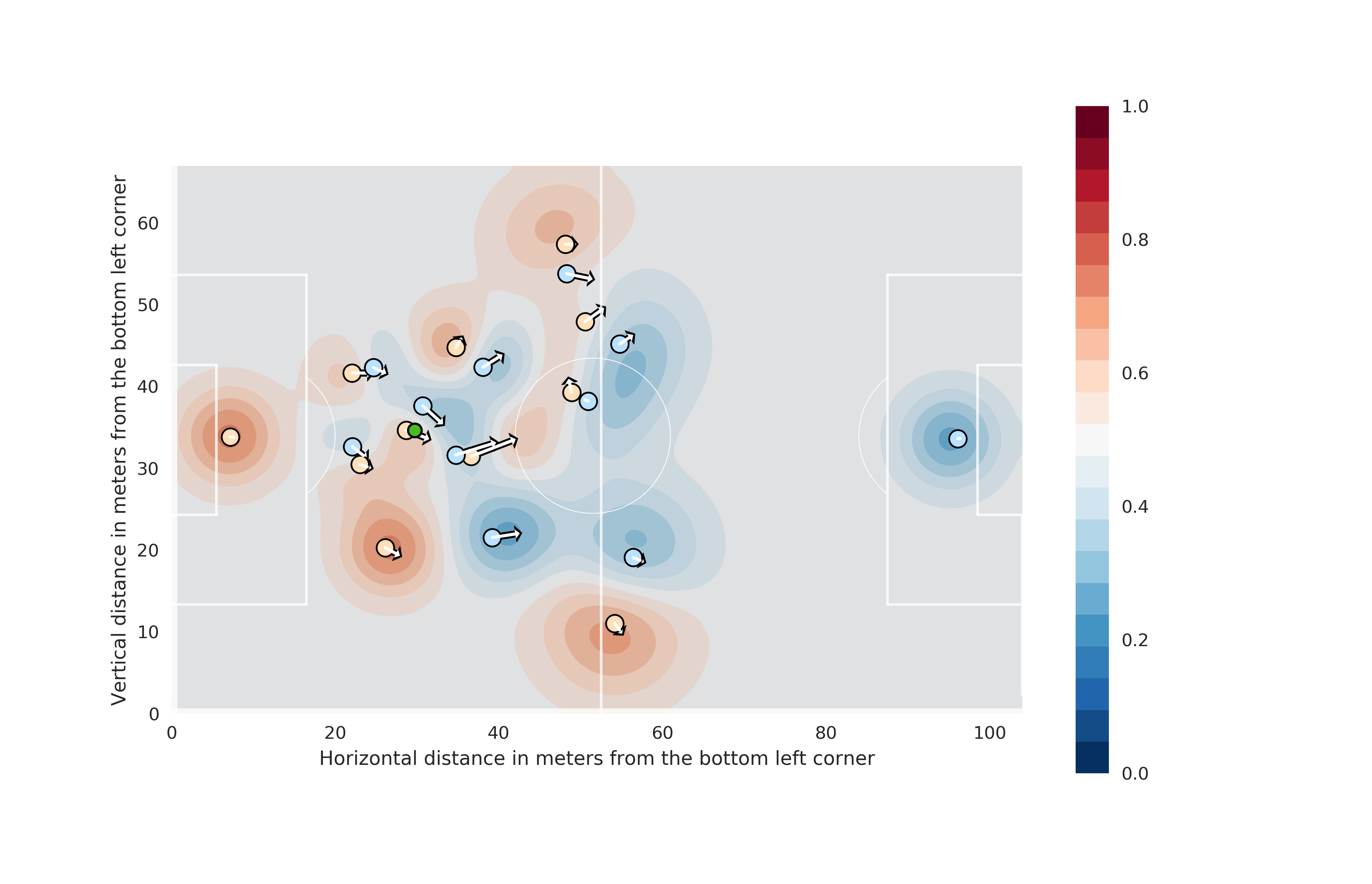}
\caption{The dots correspond to players' locations where the attacking team's players appear in blue and the opponent team's players in red. The ball location is represented as a green circle. White arrows show the direction of the player's velocity vector, ending at the expected location in one second. The surface corresponds to the attacking team's pitch control.}
\label{fig:pitch_control}       
\end{figure*}

\begin{figure*}[h!]
  \includegraphics[width=1\textwidth]{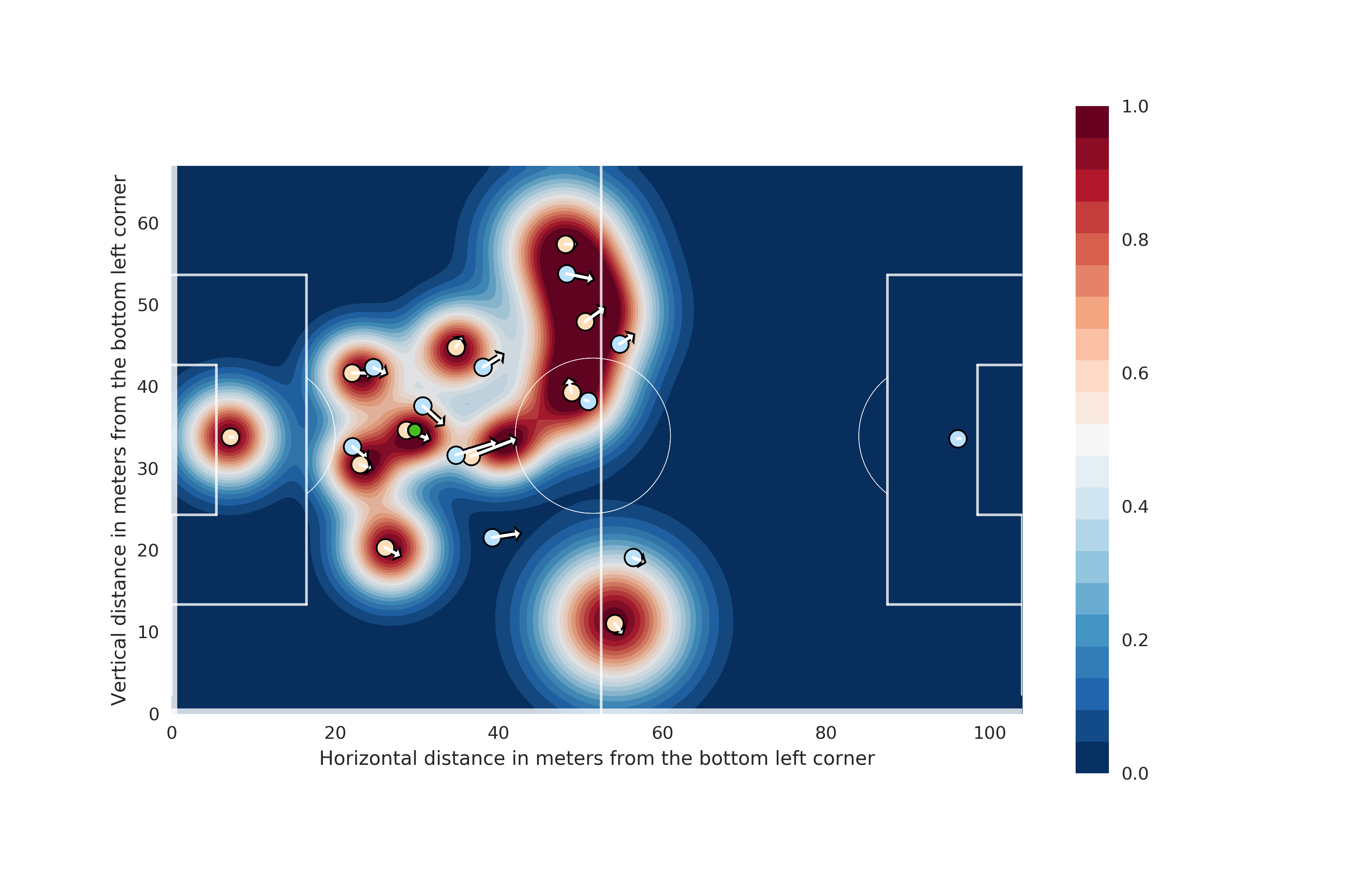}
\caption{The sum of player pitch influence in every location for every player in the attacking team. The dots correspond to players' locations where the attacking team's players appear in blue and the opponent team's players in red. The ball location is represented as a green circle. White arrows show the direction of the player's velocity vector, ending at the expected location in one second. }
\label{fig:pitch_influence}       
\end{figure*}

This influence value is normalized in the [0,1] range and provides a degree of influence for a given player. Having a quantification of individual players influence, we can calculate pitch control of a team $PC$ as the difference between the added influence of the possession team's players and the influence of the opponent team's players, at any given location, as shown in equation \ref{eq:pitch_control}

\begin{equation}\label{eq:pitch_control}
PC(p,t) = \sigma(\gamma(\lambda_1\sum_i I_i(p,t) - \lambda_2\sum_j I_j(p,t)))
\end{equation}

where $\sigma$ is the logistic function, $\lambda_1$ and $\lambda_2$ are weight parameters to allow balancing each team's overall influence, and $\gamma$ is a shrinking factor for the input logistic function. Figure \ref{fig:pitch_control} presents this probabilistic pitch control surface on a given soccer situation, while Figure \ref{fig:pitch_influence} presents the influence surface of the attacking team players.
Pitch influence and pitch control provide a rich summary of players' spatial distribution and impact along the playing surface and can be used to enrich the information on locations where players are not directly present but are having a certain influence from soccer's tactical perspective. In this work we set these parameters to the following values $\lambda_1=1$, $\lambda_2=1$, and $\gamma=1$.

\section{Dynamic pressure lines model}\label{app:dynamic_pressure_lines}

As described in Section \ref{sec:contextual_features}, we consider a contextual factor identifying the team player's alignment in a given time instance. The alignment of players is often observed in soccer through concepts such as team formation or the identification of forwards, midfielders, and defenders. However, this organization of players is manifested dynamically during the game. Instead of following a strict and predefined positioning, players tend to adapt their location to the specific situation of a given time instance in the game. Specifically, while defending, players tend to align within groups of pressure across the field. We call this alignment group dynamic pressure lines.\\

Extending from this idea, we first define dynamic pressure lines with higher generality as the centroids of a number $k$ of clusters representing hard partitions for players of the same team, where the intra-cluster distance is minimized, and the inter-cluster distance is maximized. If the clustering is based on the breadth-wise location of players (x-axis), we call them vertical dynamic pressure lines, and if it is based on the depth-wise location of players (y-axis), we call them horizontal dynamic pressure lines. 

\begin{definition}\label{def:dynamic_lines}
Given a set of $n$ player locations $ P = \{p_1,...,p_n\}$, and let $d(p,q)$ be the Euclidean distance between $p$ and $q$, and $D(L_1,L_2)$ the distance between clusters $L_1$ and $L_2$, the set $L$ of dynamic pressure lines is conformed by the average locations of the player's belonging to the complete-linkage clustering of $P$ in $k$ partitions, such that for $L_1,L_2 \in L$  and $D(L_1,L_2) = \max_{p^{L_1},p^{L_2}} d(p^{L_1},p^{L_2})$. When $p_i = (x_i,y_i) = (x_i,0)$ we  call $L$ the set of vertical dynamic pressure lines, and when $p_i = (x_i,y_i) = (0,y_i)$ we call $L$ the set of horizontal dynamic pressure lines.
\end{definition}

In this work, we set $k=3$ to identify vertical pressure lines, which conceptually represent forwards, midfielders, and defenders. For horizontal pressure lines, we set $k=3$, which will tend to define the breadth-wise borderlines of the team formation block and split the inside of the block into two parts.  

\section{SoccerMap architecture}\label{app:soccernet}
\begin{figure*}
  \includegraphics[width=1\textwidth]{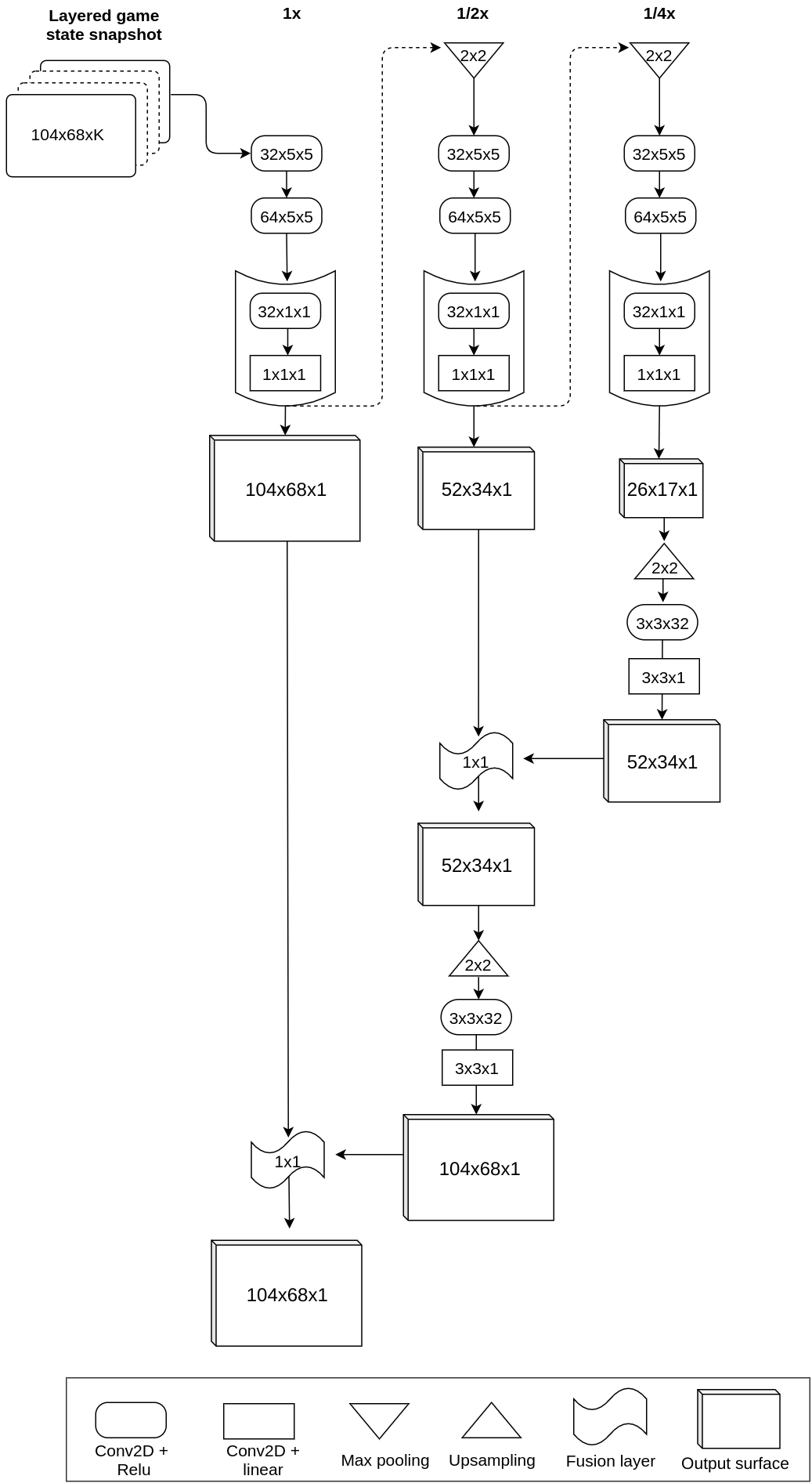}
\caption{SoccerMap architecture used as the feature extractor block component. A layered input of a game state snapshot is fed to a network which produces the input at 1x, 1/2x and 1/4x sampling scales in order to capture both local and global features. Outputs at different sampling rates are merged together and upsampled to produce a single prediction surface.}
\label{fig:feature_extractor_block}
\end{figure*}

Fully convolutional networks focus on estimating a full prediction surface from the input data, contrasting with the typical application of convolutional neural networks for classification, where outcomes tend to be either binomially or multinomially distributed. Such is the case of the image segmentation problem where, given an image, we intend to estimate a pixel-level correspondence to multiple objects present in the input image \citep{long2015fully,yu2015multi,pathak2015constrained}. 
SoccerMap is modeled as a fully convolutional network-based architecture. In its design, SoccerMap uses several components of successful architectures in other application fields such as convolutional filters, pooling and upsampling, fusion layers, and activation layers. Figure \ref{fig:feature_extractor_block} presents the standard architecture for the feature extractor block, for a soccer field representation of sizes 104$\times$68. First, the input data constituted by the layered data snapshot $\Upsilon_t$ goes through two layers of 32 and 64 convolutional filters, respectively, with 5$\times$5 activation fields and stride of 1, and then is downsampled to 1/2x using max-pooling. This process is repeated twice, so three outputs of convolutional filters are produced at 1x, 1/2x, and 1/4x sampling scales. Each of the three outputs at each scale is fed to convolutional prediction layers that produce a  prediction matrix at each sampling scale. The predictions at each scale are merged (previous upsampling to match dimensions) through convolutional layers with linear activation, called fusion layers. The output is fed to a final prediction layer constituted by 1$\times$1 convolutional filters of stride 1, producing a 104$\times$68$\times$1 surface. The combination of layers at different resolutions allows capturing relevant information at both local and global levels, with the expectation of producing location-wise predictions that are spatial-aware. This approach is inspired by a nonlinear feature hierarchy called deep jet \citep{long2015fully}.

\section{Baseline expected goals model}\label{app:xg}
\begin{table}[h!]\label{table:baseline_xg}
\caption{Total count of matches and shot events included within the event-data dataset.}
\begin{tabularx}{\textwidth}{XXXXXX} 
\hline\noalign{\smallskip}
Data Type & Source   & \# Total & \# Training  & \# Test & \% Goals  \\ 
\noalign{\smallskip}\hline\noalign{\smallskip}
Match     & Event    & 4,679        & 3,509           & 1,170    & -           \\
Shot      & Event    & 117,948      & 87,980         & 30,645   & 10.4        \\
\noalign{\smallskip}\hline
\end{tabularx}
\end{table}

In several of the models presented in this work and in the definition of the outcome of the possession presented in Section \ref{sec:estimands}, we use a general estimation of the goal expectation given a shot is taken, based on event-data. 
In order to produce a calibrated baseline estimation of expected goals, we use a wide dataset of event-data provided by \emph{OPTA}, which contains $117,948$ shot events and $12,266$ goals as detailed in Table \ref{table:baseline_xg}. Despite only providing the location and time of observed shots, this dataset is considerably larger than the $13,735$ shots available in the tracking data dataset (see Section \ref{sec:datasets}). Event-data has been used successfully in previous work to obtain a calibrated estimation of expected goals \citep{eggels2016expected}. \\

We use a set of spatial features by the event location and the distance and angle between the ball location and the goal. Contextual features are composed of a one-hot encoded vector indicating the attacking type at the moment of the event (regular-play, set-piece, free-kick, corner, penalty), and a boolean variable indicating whether the action is taken with the head or not. The matches are split into a training and test set. For every shot in the dataset, we label the outcome as 1 of the shots results in a goal, and 0 otherwise. We build the model using the extreme gradient boosting algorithm XGBoost \citep{chen2016xgboost}, and we perform an exhaustive grid-search on the following hyper-parameters of the model: number of trees  ($\{50,100,250\}$), learning rate ($\{1e-3,1e-2,1e-1\}$), and maximum depth ($\{3,5,10\}$). Model selection is performed through a K-fold cross-validation procedure on the training set, with $K=10$. All the features are standardized, obtaining a scaled feature set where each variable has a mean of $0$ and a unitary standard deviation.\\

The best model presented a log loss value of $0.2540$ and a calibration ECE value of $0.00594$, in the test set. The parameters of the best model where: $100$ trees, a maximum depth of $3$, and a learning rate of $1e-1$.


\newpage
\bibliographystyle{spbasic}
\bibliography{bibliography} 

\begin{thebibliography}{22}
\providecommand{\natexlab}[1]{#1}
\providecommand{\url}[1]{{#1}}
\providecommand{\urlprefix}{URL }
\expandafter\ifx\csname urlstyle\endcsname\relax
  \providecommand{\doi}[1]{DOI~\discretionary{}{}{}#1}\else
  \providecommand{\doi}{DOI~\discretionary{}{}{}\begingroup
  \urlstyle{rm}\Url}\fi
\providecommand{\eprint}[2][]{\url{#2}}

\bibitem[{Bransen and Van~Haaren(2018)}]{bransen2018measuring}
Bransen L, Van~Haaren J (2018) Measuring football players’ on-the-ball
  contributions from passes during games pp 3--15

\bibitem[{Cervone et~al.(2016)Cervone, D'Amour, Bornn, and
  Goldsberry}]{cervone2016multiresolution}
Cervone D, D'Amour A, Bornn L, Goldsberry K (2016) A multiresolution stochastic
  process model for predicting basketball possession outcomes. Journal of the
  American Statistical Association 111(514):585--599

\bibitem[{Chen and Guestrin(2016)}]{chen2016xgboost}
Chen T, Guestrin C (2016) Xgboost: A scalable tree boosting system. In:
  Proceedings of the 22nd acm sigkdd international conference on knowledge
  discovery and data mining, ACM, pp 785--794

\bibitem[{Decroos et~al.(2019)Decroos, Bransen, Van~Haaren, and
  Davis}]{decroos2018actions}
Decroos T, Bransen L, Van~Haaren J, Davis J (2019) Actions speak louder than
  goals: Valuing player actions in soccer. In: Proceedings of the 25th ACM
  SIGKDD International Conference on Knowledge Discovery \& Data Mining, pp
  1851--1861

\bibitem[{Eggels(2016)}]{eggels2016expected}
Eggels H (2016) Expected goals in soccer: Explaining match results using
  predictive analytics. In: The Machine Learning and Data Mining for Sports
  Analytics workshop, p~16

\bibitem[{Fernandez and Bornn(2018)}]{fernandez2018wide}
Fernandez J, Bornn L (2018) Wide open spaces: A statistical technique for
  measuring space creation in professional soccer. In: Sloan Sports Analytics
  Conference

\bibitem[{Fern{\'a}ndez and Bornn(2020)}]{fernandez2020soccermap}
Fern{\'a}ndez J, Bornn L (2020) Soccermap: A deep learning architecture for
  visually-interpretable analysis in soccer. arXiv preprint arXiv:201010202

\bibitem[{Guo et~al.(2017)Guo, Pleiss, Sun, and
  Weinberger}]{guo2017calibration}
Guo C, Pleiss G, Sun Y, Weinberger KQ (2017) On calibration of modern neural
  networks. In: Proceedings of the 34th International Conference on Machine
  Learning-Volume 70, JMLR. org, pp 1321--1330

\bibitem[{Gyarmati and Stanojevic(2016)}]{gyarmati2016qpass}
Gyarmati L, Stanojevic R (2016) Qpass: a merit-based evaluation of soccer
  passes. arXiv preprint arXiv:160803532

\bibitem[{Hub{\'a}{\v{c}}ek et~al.(2018)Hub{\'a}{\v{c}}ek, {\v{S}}ourek, and
  {\v{Z}}elezn{\`y}}]{hubavcek2018deep}
Hub{\'a}{\v{c}}ek O, {\v{S}}ourek G, {\v{Z}}elezn{\`y} F (2018) Deep learning
  from spatial relations for soccer pass prediction. In: International Workshop
  on Machine Learning and Data Mining for Sports Analytics, Springer, pp
  159--166

\bibitem[{Kingma and Ba(2014)}]{kingma2014adam}
Kingma DP, Ba J (2014) Adam: A method for stochastic optimization. arXiv
  preprint arXiv:14126980

\bibitem[{Link et~al.(2016)Link, Lang, and Seidenschwarz}]{link2016real}
Link D, Lang S, Seidenschwarz P (2016) Real time quantification of dangerousity
  in football using spatiotemporal tracking data. PloS one 11(12):e0168768

\bibitem[{Liu and Schulte(2018)}]{liu2018deep}
Liu G, Schulte O (2018) Deep reinforcement learning in ice hockey for
  context-aware player evaluation. arXiv preprint arXiv:180511088

\bibitem[{Long et~al.(2015)Long, Shelhamer, and Darrell}]{long2015fully}
Long J, Shelhamer E, Darrell T (2015) Fully convolutional networks for semantic
  segmentation. In: Proceedings of the IEEE conference on computer vision and
  pattern recognition, pp 3431--3440

\bibitem[{Lucey et~al.(2014)Lucey, Bialkowski, Monfort, Carr, and
  Matthews}]{lucey2014quality}
Lucey P, Bialkowski A, Monfort M, Carr P, Matthews I (2014) quality vs
  quantity: Improved shot prediction in soccer using strategic features from
  spatiotemporal data. In: Proc. 8th annual mit sloan sports analytics
  conference, pp 1--9

\bibitem[{Pathak et~al.(2015)Pathak, Krahenbuhl, and
  Darrell}]{pathak2015constrained}
Pathak D, Krahenbuhl P, Darrell T (2015) Constrained convolutional neural
  networks for weakly supervised segmentation. In: Proceedings of the IEEE
  international conference on computer vision, pp 1796--1804

\bibitem[{Power et~al.(2017)Power, Ruiz, Wei, and Lucey}]{power2017not}
Power P, Ruiz H, Wei X, Lucey P (2017) Not all passes are created equal:
  Objectively measuring the risk and reward of passes in soccer from tracking
  data. In: Proceedings of the 23rd ACM SIGKDD International Conference on
  Knowledge Discovery and Data Mining, ACM, pp 1605--1613

\bibitem[{Rudd(2011)}]{rudd2011framework}
Rudd S (2011) A framework for tactical analysis and individual offensive
  production assessment in soccer using markov chains. In: New England
  Symposium on Statistics in Sports. http://nessis.org/nessis11/rudd.pdf

\bibitem[{Singh(2019)}]{xtkarun}
Singh K (2019) Introducing expected threat (xt).
  \url{https://karun.in/blog/expected-threat.html}, accessed: 2020-10-16

\bibitem[{Spearman(2018)}]{spearmanbeyond}
Spearman W (2018) Beyond expected goals. In: Proceeding of the 12th MIT Sloan
  Sports Analytics Conference

\bibitem[{Yu and Koltun(2015)}]{yu2015multi}
Yu F, Koltun V (2015) Multi-scale context aggregation by dilated convolutions.
  arXiv preprint arXiv:151107122

\bibitem[{Yurko et~al.(2020)Yurko, Matano, Richardson, Granered, Pospisil,
  Pelechrinis, and Ventura}]{yurko2019going}
Yurko R, Matano F, Richardson LF, Granered N, Pospisil T, Pelechrinis K,
  Ventura SL (2020) Going deep: models for continuous-time within-play
  valuation of game outcomes in american football with tracking data. Journal
  of Quantitative Analysis in Sports 1(ahead-of-print)

\end{thebibliography}

\end{document}